\documentclass[authorversion, sigconf]{acmart}
\usepackage{subfig}

\usepackage{xcolor}
\usepackage{multirow}
\usepackage[normalem]{ulem}
\usepackage{amsmath}
\usepackage{xspace}
\usepackage{soul}
\usepackage{placeins}

\usepackage{subfig}
\usepackage{url}
\usepackage{etex}
\usepackage{etoolbox}
\usepackage{xspace}
\usepackage{amsmath}
\usepackage{xcolor}
\usepackage{nicefrac}
\usepackage{changepage}
\usepackage{array,booktabs,multirow}
\usepackage{afterpage}
\usepackage{pbox}
\RequirePackage{enumerate}
\RequirePackage{paralist}
\RequirePackage{enumitem}
\usepackage{paralist}
\usepackage{mdwlist}

\usepackage{hhline}
\usepackage{lipsum} 
\usepackage{adjustbox}
\usepackage{tabu}
\usepackage{makecell}
\usepackage{color, colortbl}
\usepackage{amsmath,amssymb,latexsym}

\usepackage{algorithm}
\usepackage[noend]{algpseudocode}

\usepackage{bm}
\usepackage[super]{nth}

\usepackage{amsfonts}

\usepackage{pifont}

\usepackage{xargs}                                                                        
\usepackage[colorinlistoftodos,prependcaption,textsize=small]{todonotes}                  
\newcommandx{\unsure}[2][1=]{\todo[linecolor=red,backgroundcolor=red!25,bordercolor=red,#1]{#2}}
\newcommandx{\change}[2][1=]{\todo[linecolor=blue,backgroundcolor=blue!25,bordercolor=blue,#1]{#2}}
\newcommandx{\info}[2][1=]{\todo[linecolor=OliveGreen,backgroundcolor=OliveGreen!25,bordercolor=OliveGreen,#1]{#2}}
\newcommandx{\improvement}[2][1=]{\todo[linecolor=Plum,backgroundcolor=Plum!25,bordercolor=Plum,#1]{#2}}
\newcommandx{\thiswillnotshow}[2][1=]{\todo[disable,#1]{#2}}   

\makeatletter                                                                             
\newcommand{\thickhline}{%
    \noalign {\ifnum 0=`}\fi \hrule height 1.2pt                                          
    \futurelet \reserved@a \@xhline                                                       
}                                                                                         
\makeatletter                                                                             
\newcommand{\midhline}{%
    \noalign {\ifnum 0=`}\fi \hrule height 1pt                                            
    \futurelet \reserved@a \@xhline                                                       
}    

\makeatletter
\def\SOUL@hlpreamble{%
    \setul{0ex}{2ex}
    \let\SOUL@stcolor\SOUL@hlcolor
    \SOUL@stpreamble
}

\usepackage{algorithm}
\usepackage[noend]{algpseudocode}

\floatevery{algorithm}{\setlength\hsize{3.7cm}}

\algnewcommand{\LineComment}[1]{\State \(\triangleright\) #1}     
\makeatletter
\renewcommand{\ALG@beginalgorithmic}{\small}
\makeatother                        
                                                                                          
\algblockdefx{FORP}{ENDFP}[1]%
{\textbf{for all }#1 \textbf{do in parallel}}%
{\textbf{end for}}

\algblockdefx{FORB}{ENDFB}[1]%
{\textbf{for }#1 \textbf{do in backward}}%
{\textbf{end for}}

\makeatletter                                                                             
\newcommand{\algorithmfootnote}[2][\footnotesize]{%
    \let\old@algocf@finish\@algocf@finish
    \def\@algocf@finish{\old@algocf@finish
        \leavevmode\rlap{\begin{minipage}{\linewidth}                                     
                #1#2                                                                      
        \end{minipage}}%
    }%

}             

\makeatletter  
\def\@eqnnum{{\normalfont\normalcolor[\theequation]}}  
\makeatother

\newcommand{\ceil}[1]{\lceil #1 \rceil}

\newcommand{\abbrfnt}{}
\newcommand{\abbrev}[1]{{\mbox{\abbrfnt{#1}}}\xspace}
\newcommand{\sect}[1]{Section.~\ref{#1}\xspace}
\newcommand{\algo}[1]{Algorithm.~\ref{#1}\xspace}
\newcommand{\eq}[1]{Eq.~\ref{#1}\xspace}

\makeatletter
\renewcommand{\fnum@table}{Tab. \thetable}
\makeatother
\newcommand{\tab}[1]{Tab.~\ref{#1}\xspace}

\makeatletter
\renewcommand{\fnum@figure}{Fig. \thefigure}
\makeatother
\newcommand{\fig}[2][]{Fig.~\ref{#2}#1\xspace}

\newcommand{\squeezetrain}{\abbrev{PruneTrain}}
\newcommand{\scratch}[1]{}

\newcommand{\mdel}[2]{}

\newcommand{\FIXME}[1]{}

\begin{document}
\title{\squeezetrain: Fast Neural Network Training by Dynamic Sparse Model Reconfiguration}

\author{Sangkug Lym}
\affiliation{\institution{The University of Texas at Austin}}
\email{sklym@utexas.edu}

\author{Esha Choukse}
\affiliation{\institution{The University of Texas at Austin}}
\email{esha.choukse@utexas.edu}

\author{Siavash Zangeneh}
\affiliation{\institution{The University of Texas at Austin}}
\email{siavash.zangeneh@utexas.edu}

\author{Wei Wen}
\affiliation{\institution{Duke University}}
\email{wei.wen@duke.edu}

\author{Sujay Sanghavi}
\affiliation{\institution{The University of Texas at Austin}}
\email{sanghavi@mail.utexas.edu}

\author{Mattan Erez}
\affiliation{\institution{The University of Texas at Austin}}
\email{mattan.erez@utexas.edu}

\begin{abstract}
  State-of-the-art convolutional neural networks (CNNs) used in vision applications have large models with numerous weights. Training these models is very compute- and memory-resource intensive. Much research has been done on pruning or compressing these models to reduce the cost of inference, but little work has addressed the costs of training. We focus precisely on accelerating training. We propose \squeezetrain, a cost-efficient mechanism that gradually reduces the training cost during training. \squeezetrain uses a structured group-lasso regularization approach that drives the training optimization toward both high accuracy and small weight values. Small weights can then be periodically removed by reconfiguring the network model to a smaller one. By using a structured-pruning approach and additional reconfiguration techniques we introduce, the pruned model can still be efficiently processed on a GPU accelerator. Overall, \squeezetrain achieves a reduction of 39\% in the end-to-end training time of ResNet50 for ImageNet by reducing computation cost by 40\% in FLOPs, memory accesses by 37\% for memory bandwidth bound layers, and the inter-accelerator communication by 55\%.

\end{abstract}

\copyrightyear{2019} 
\acmYear{2019} 
\acmConference[SC '19]{The International Conference for High Performance Computing, Networking, Storage, and Analysis}{November 17--22, 2019}{Denver, CO, USA}
\acmBooktitle{The International Conference for High Performance Computing, Networking, Storage, and Analysis (SC '19), November 17--22, 2019, Denver, CO, USA}
\acmPrice{15.00}
\acmDOI{10.1145/3295500.3356156}
\acmISBN{978-1-4503-6229-0/19/11}

\maketitle
\section{Introduction}
\label{sec:intro}

Training a modern convolutional neural network (CNN) requires millions of computation and memory bandwidth-intensive iterations. In addition, ever-growing network complexity and training dataset sizes are making the already expensive CNN training even more costly.
To accelerate the training of complex modern CNNs, a cluster of accelerators is typically used~\cite{goyal2017accurate,you2018imagenet}. However, training such complex networks on a large dataset, e.g., ImageNet (ILSVRC) ~\cite{imagenet_cvpr09}, is still a challenging problem.
To reduce this high complexity of training and eventually the training time, we use model pruning.
Model pruning involves reducing the number of learning parameters (or weights) in an initially-dense network leading to lower memory and inference costs while losing the accuracy of the original dense model as little as possible~\cite{han2015deep}.
Although the main goal of model pruning is improving the performance of inference, we find that it can also substantially accelerate training by reducing its computation, memory, and communication costs.
Our technique speeds up the time needed to train a pruned ResNet50 on ImageNet by up to 39\%, which eventually generates a dense pruned model with 47\% less inference FLOPs and 1.9\% lower accuracy.

Numerous model pruning mechanisms have been proposed for high-performance and energy-efficient inference.
They either individually remove less important parameters (with small values)~\cite{han2015deep,han2015learning}, or structurally remove a group of such parameters~\cite{NIPS2016_6504,feng2015learning,alvarez2017compression,he2017channel,he2018amc}.
Most such prior work performs model pruning using a pre-trained model, and actually increases end-to-end training time as a result.

A few other prior works prune the model during training. Training is an optimization process to minimize the loss function, which represents the error (typically cross-correlation) between predictions and the training-set ground truth.
One approach to prune during training is to add regularization terms to the loss function, such as the $l_1$-$norm$ of weights or of a group lasso~\mbox{\cite{yuan2006model}} that uses $l_1$-$norms$ or $l_2$-$norms$ of groups of weights for structural pruning. This causes the optimization process to prefer small absolute values for weights or groups of weights, sparsifying the model.
Very small weights and their associated momentum and normalization parameters can then be zeroed out, or \emph{pruned}.

Although these prior works can prune during training, they do not speed up training effectively. Most prior works maintain the original dense CNN structure even after pruning and thus cannot save computation, while others require complex and performance-reducing data indexing to process a sparse CNN structure~\cite{NIPS2016_6504,wen2017learning,zhou2016less}.
The closest prior work to ours reconfigures the CNN architecture exactly once during the training process, processing the smaller model from that point on~\cite{alvarez2017compression}.
However, the reconfiguration point is not known a priori, making applying this approach problematic, or even counterproductive.

\begin{figure}[t]
    \centering
    \includegraphics[width=0.48\textwidth]{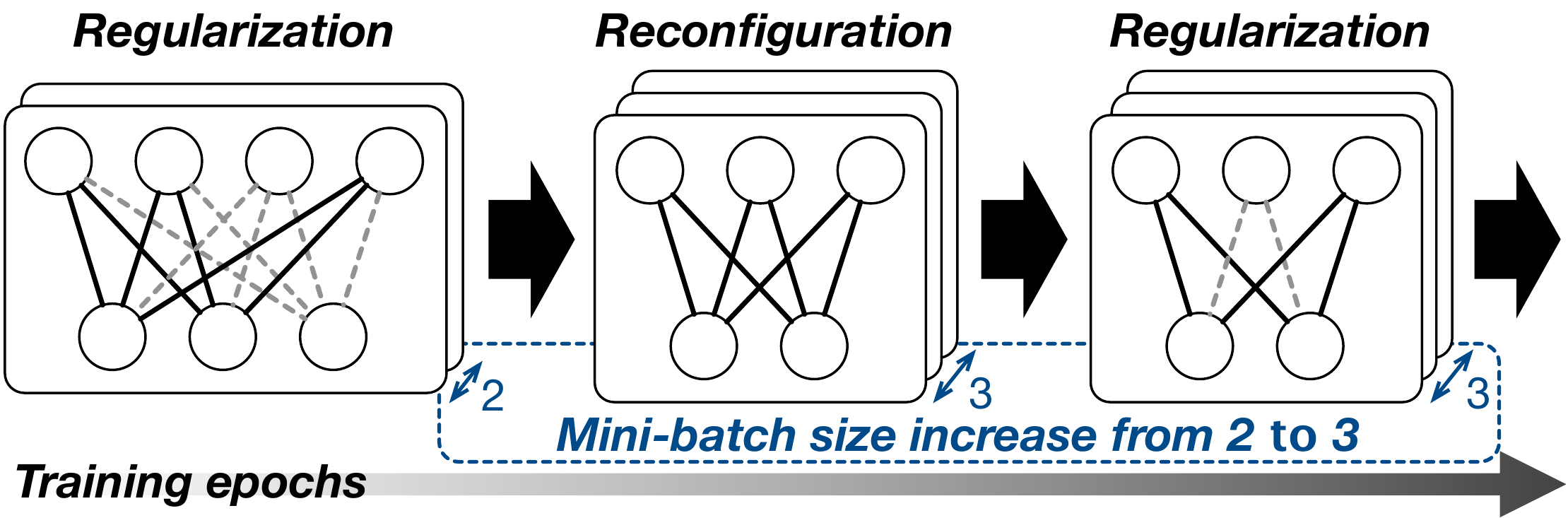}
    \vspace*{-4mm}
    \caption{\squeezetrain process: weights of each channel are continuously regularized to small absolute values during training. The sparsified channels, whose all input or output connections become dotted, are removed and the network architecture is reconfigured into a new dense form. Decreasing training memory capacity requirement after each reconfiguration enables using a larger mini-batch.}
    \label{fig:main}
\end{figure}

We propose {\bf \squeezetrain}, a CNN training acceleration mechanism that, unlike prior work, prunes the model during training from scratch with the \emph{sparsification process} starting during the first training epoch. We use \emph{group lasso regularization}~\cite{meier2008group} as the baseline sparsification approach and then periodically prune weights and reconfigure the CNN to continue training on the pruned model. For efficient execution on data-parallel training accelerators (e.g., GPUs), we group parameters at channel granularity and prune those channels for which all parameters are below a threshold. As a result, the periodic reconfiguration maintains a still dense, yet smaller model (\fig{fig:main}). This model, which requires less computation, memory, and communication, continues to shrink as sparsification and pruning continue throughout training. This approach is possible because, as we observe, once weights are sparsified by group lasso, they rarely grow to above threshold later in training; these sparsified weights almost never \emph{revive} and can be pruned without degrading accuracy. \squeezetrain reduces the computations of ResNet50 for ImageNet, a most commonly used modern image classifier, by 40\%, the memory traffic of memory bound layers (e.g. batch normalization) by 37\%, and the inter-GPU communication cost by 55\% compared to the dense baseline training.

For the efficient realization of \squeezetrain, we introduce three key optimization techniques. First, we propose a {\bf systematic method to set the group lasso regularization penalty coefficient} at the beginning of training. This penalty coefficient is a hyperparameter that trades off model size with accuracy. Prior work searches for an appropriate penalty value, making it expensive to include pruning from the beginning of training.  Our mechanism effectively controls this group lasso regularization strength and achieves a high model pruning rate with small impact on accuracy with even a single training run.

Second, we introduce {\bf channel union}, a memory-access cost-efficient and index-free channel pruning algorithm for modern CNNs with short-cut connections. Short-cut connections (e.g., residual blocks in ResNet~\cite{he2016deep}) are widely used in modern CNNs~\cite{huang2017densely,hu2018squeeze,zagoruyko2016wide}. Pruning all the zeroed channels of such CNNs require frequent tensor reshaping to match channel indices between layers. Such reshaping or indexing decreases performance. Our channel union algorithm does not require any zeroed channel indexing and tensor reshaping, and can thus accelerate convolution layer performance by 1.9X on average compared to a dense baseline; if indexing is used, training is slowed down rather than accelerated.

Lastly, we propose {\bf dynamic mini-batch adjustment} that dynamically adjusts the size of the mini-batch (the number of samples used for each stochastic gradient descent step) by monitoring the memory capacity requirement of a training iteration after each pruning reconfiguration (\fig{fig:main}). Dynamic mini-batch adjustment compensates for the reduced data parallelism of the smaller pruned model by increasing the mini-batch size. This both improves HW resource utilization in processing a pruned model and reduces the communication overhead by decreasing the model update frequency. When increasing the mini-batch size, our algorithm increases the learning rate by the same ratio to avoid affecting accuracy~\cite{smith2017don}.

We summarize our contributions as follows:
\begin{itemize}[noitemsep,topsep=5pt,leftmargin=0.15in]
	\item 
        We propose \emph{\squeezetrain} to continuously prune a CNN model and reconfigure its architecture into a more cost-efficient but still dense form. \squeezetrain accelerates model training by reducing computation, memory access, and communication costs.
    \item
        We propose a systematic method to set the regularization penalty coefficient that enables parameter regularization from the beginning of training and achieves high model pruning rate with minor accuracy loss by a single training run.
    \item
        We propose \emph{channel union} that does not require any complex channel indexing or tensor reshaping of processing a pruned CNN model with short-cut connections by negligible computation cost increase.
	\item
	    We dynamically increases the mini-batch size by monitoring the memory capacity requirement of a training iteration, which increases the data parallelism and reduces inter-accelerator communication frequency leading to shorter training time compared to the baseline \squeezetrain.
\end{itemize}


\section{Background and Related Work}
\label{sec:background}

Although the proposed mechanisms of \squeezetrain are applicable to different neural networks, e.g., recurrent neural networks, we describe \squeezetrain in the context of CNNs in this paper.

\subsection{CNN Architecture}
\label{sec:cnn_arch}
A CNN consists of various layer operators and the performance of different layer types is bounded by different HW resources. Convolution and feature normalization layers account for the majority of the training time of modern CNNs. Convolution layers extract features from input images for pattern recognition and their computation consists primarily of matrix multiplication and accumulation. Since convolution layers exhibit high input and output data locality, their execution time is bounded by the computation throughput of an accelerator.

Feature normalization layers (e.g batch normalization) maintain stable feature distribution across layers and different input samples~\cite{ioffe2015batch} to enable a deep layer architecture and fast training convergence. Normalization layers read their inputs multiple times to calculate mean, variance, and normalize them, which typically takes $\sim$30\% of CNN training time~\cite{hoffer2018norm}. Due to low arithmetic intensity, the performance of normalization layers is bounded by memory access bandwidth. CNNs also contain other types of layers such as a fully connected (FC) layers that generate class scores, down-sampling layers that reduce the size of features, and many other element-wise operation layers. However, their execution time in training is relatively negligible.

\subsection{CNN Model Training}
\label{sec:cnn_train}
The weights (or model parameters) of a CNN are trained in three steps. First, a network takes in input samples, forward propagates them through layers, and attempts to predict the correct outputs using its current weights. Then, it compares its prediction outputs to their ground truth and computes an average loss (or prediction error). Next, this loss is back-propagated through layers and, at each layer, gradients of the loss w.r.t. the weights are calculated. Finally, the original weights are updated by using these weight gradients and an optimization algorithm~\cite{ruder2016overview}.

Mini-batch SGD (stochastic gradient descent) is the most commonly used CNN training algorithm, which uses a set of input samples for each training iteration. Using a large mini-batch exhibits many benefits: (1) it provides abundant data parallelism to each layer operation, which helps achieve high HW resource utilization, (2) reduces the frequency of weight updates, and (3) decreases the variance in weight updates between training iterations~\cite{li2014efficient}.

\medskip
\noindent\textbf{Distributed Training.}
A cluster of GPUs is typically used to train a complex CNN model on a large dataset. Data parallelism~\cite{krizhevsky2012imagenet} is the most commonly used multi-processor training mechanism. First, each GPU in the system holds the same copy of weights. Then, the mini-batch of input samples are distributed to each GPU and all GPUs process the inputs in parallel. Data parallelism is network traffic-efficient as the inter-GPU communication is required only for the model updates; the partial weight gradients of all GPUs are first reduced then used to update the current weights. Although using more GPUs increases the peak computation throughput, it also increases this communication overhead, preventing linear end-to-end training performance scaling. For efficient weight gradient reduction, ring-allreduce based communication is commonly used for weight gradients reduction, which efficiently pipelines data transfer latencies among nodes~\cite{wen2017terngrad}. In particular, recently proposed hierarchical allreduce communication~\cite{li2018network} reduces the communication complexity by hierarchically dividing the reduction granularity and achieves more linear training performance scaling with increasing number of GPUs.

\medskip
\noindent\textbf{Training Memory Context.}
Processing a training iteration requires large off-chip memory space. This is mainly because the inputs of each layer at forward propagation should be kept in memory and reused to compute the local gradients in back-propagation. In particular, the total size of all layer inputs linearly increases with mini-batch size~\cite{lym2018mini}. Therefore, small off-chip memory capacity or a large feature size of a CNN can constrain the mini-batch size per accelerator, and hence also the data parallelism of each layer. This eventually decreases HW resource utilization. In addition, insufficient memory increases the total number of training iterations per epoch because of smaller mini-batches, which increases the communication cost for model updates.

\subsection{Network Model Pruning}
\label{subsec:pruning}
Model pruning has been studied primarily for CNNs, to make their models more compact and their inference fast and energy-efficient. Most pruning methods compress a CNN model by removing small-valued weights with a fine-tuning process to minimize accuracy loss~\cite{han2015deep,han2015learning}. Pruning algorithms can be unstructured or structured. Unstructured pruning can maximize model-size reduction but requires fine-grained indexing with irregular data access patterns. Such accesses and extra index operations lead to poor performance on deep learning accelerators with vector or matrix computing units despite reducing the number of weights and FLOPs~\cite{han2017ese,yu2017scalpel,anwar2017structured}. Structured-pruning algorithms remove or reduce fine-grained indexing and better match the needs of hardware and thus effectively realize performance gains.

\medskip
\noindent\textbf{Trial-and-Error Based Structured Model Pruning.}
One approach to structured pruning is to start with a pre-trained dense model and then attempt to remove weights in a structured manner, generally removing channels rather than individual weights~\cite{he2017channel,hu2016network,molchanov2016pruning,he2018amc}. Unimportant channels are removed based on the value of their weights or hints derived from regression~\cite{tibshirani1996regression}. The removed channels are rolled back if accuracy is severely affected. Although effective, the search space of such a trial-and-error based model pruning substantially increases with the complexity of the network model, which can increase pruning time significantly. Also, as pruning is applied to a pre-trained model, these mechanisms do not speed up training.

\medskip
\noindent\textbf{Related Work: Structured Pruning During Training.}
An alternative mechanism to trial-and-error pruning uses \emph{parameter regularization}. This optimizes training loss while simultaneously forcing the absolute values of weights or groups of weights toward zero. We call this process of forcing weights toward zero \emph{sparsificiation}. 
Group lasso regularization is typically used to structurally sparsify weights by assigning a regularization penalty to $l_2$-$norms$ of groups of weights~\cite{NIPS2016_6504,wen2017learning,feng2015learning,alvarez2017compression,zhou2016less}.

This regularization-based pruning mechanism adds regularization loss terms to the baseline classification loss function, then back-propagate the loss to update the weights to both improve accuracy and reduce their absolute values. Eventually, the sparsified weights can be effectively zeroed-out and pruned from the model.

In particular, Wen et al.~\cite{NIPS2016_6504} propose \emph{SSL}, a pruning mechanism that sparsify weights while training a CNN. However, they start from a pre-trained model and maintain the original dense network architecture until the end of training because sparsified weights may \emph{revive} later in training. Thus, SSL actually requires more time to train, first training the dense baseline and then pruning it. The pruning mechanism proposed by Zhou et al.~\cite{zhou2016less} prunes the zeroed parameters during training but does not reconfigure the network architecture. Instead, gradient updates in back-propagation are skipped by setting weight momentum to zero. Since this mechanism still performs all training computation, no training performance improvement is achieved.

On the other hand, Alvarez and Salzmann~\cite{alvarez2017compression} propose to reconfigure the sparse network architecture and reload the model to accelerate training. However, they reconstruct the network only once at specific training epoch. This misses the opportunity to further improve training performance by timely network reconfiguration especially because a good reconfiguration point is not known a priori.


\section{motivation for continuous pruning and reconfiguration}
\label{sec:motivation}

\begin{figure}[t!]
    \centering
    \includegraphics[width=0.48\textwidth]{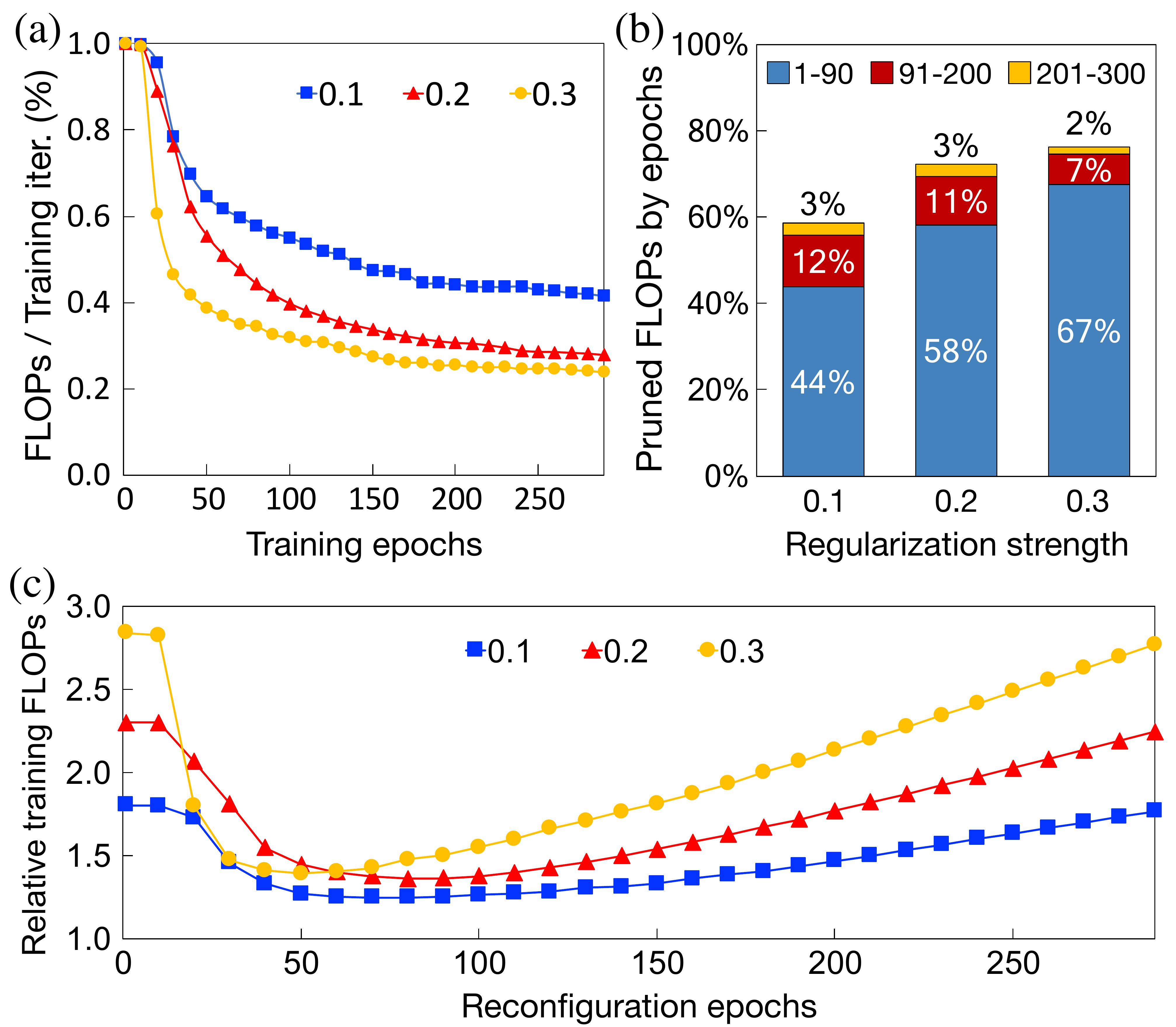}
    \vspace*{-4mm}
    \caption{(a) FLOPs per training iteration normalized to the dense baseline (ResNet50 on CIFAR10). (b) Breakdown of prunable training FLOPs over epochs. (c) Training computation overhead of one-time network reconfiguration at different training epoch compared to \squeezetrain; each line in \((a)\) and \((c)\) is the result of different sparsification strengths.}
    \label{fig:prune_rate}
    \vspace*{-1mm}
\end{figure}

Continuous pruning and reconfiguration can significantly speed up training for two reasons. First, of all the convolutional channels that regularization sparsifies, most are sparsified very early in the training process, so pruning these channels has a significant positive impact on the overall training time. Second, regularization sparsifies the channels gradually over time, so it is more beneficial to prune the sparsified channels frequently, as opposed to pruning them only once. To show this, we train ResNet50, one of the most commonly used image classifiers for various vision applications~\cite{lin2017focal,he2017mask,lin2017feature}, on the CIFAR10 dataset with regularization. Every epoch, we measure the FLOPs (floating-point operations) per training iteration, assuming we can prune the unnecessary channels every 10 epochs. \fig{fig:prune_rate}{a} shows the FLOPS per iteration normalized to the dense baseline. Each line in the figure shows the FLOPs using a different regularization strength. We will describe our definition for regularization strength in \sect{subsec:pruning_mechanism}. Regardless of the strength, the majority of FLOPs is pruned in the early epochs, with the rate of pruning gradually saturating. This is further shown by the breakdown of aggregated pruned FLOPs (\fig{fig:prune_rate}{b}) over three training phases, where most FLOPs are pruned within the first 90 training epochs.

Given that weights are gradually sparsified during training, it is apparent that continuous and timely model pruning and reconfiguration can reduce training computations much more effectively than the one-time reconfiguration proposed by Alvarez and Salzmann~\cite{alvarez2017compression}. \fig{fig:prune_rate}{c} compares the training FLOPs of one-time pruning and reconfiguration used in prior work to \squeezetrain. Regardless of the strength of group lasso regularization, even with the optimistic assumption that we know the best reconfiguration point, prior works uses more than 25\% additional training FLOPs compared to PruneTrain. In reality, it is impossible to know the best reconfiguration time a priori, and thus, PruneTrain prunes and reconfigures the models periodically.


\section{\squeezetrain}
\label{sec:squeezetrain}

We first explain the baseline group lasso regularization pruning approach also used by prior work, then describe how we modify this technique to better accelerate training and enable pruning from the first training iteration, motivate and explain our approach to dynamic reconfiguration, and finally discuss the potential for dynamically adjusting the mini-batch size as the model shrinks through pruning.

\subsection{Model Pruning Mechanism}
\label{subsec:pruning_mechanism}

\noindent\textbf{Baseline Pruning Mechanism.}
Like prior work~\cite{NIPS2016_6504,wen2017learning,alvarez2017compression}, we use group lasso regularization to sparsify weights so that they can be pruned. Group lasso regularization is a good match for PruneTrain because it is incorporated with the training optimization and imposes structure on the pruned weights, which we use to maintain an overall dense computation. Group lasso regularization modifies the optimization loss function to also include consideration for weight magnitude. This is shown in \eq{eq:optimization}, where the left term is the standard cross entropy classification loss and the right term is the general form of the group lasso regularizer. Here \(f\) is the network's prediction on the input \(x_i\), \(\bm{W}\) are the weights, \(l\) is the classification loss function between the prediction and its ground truth \(y_i\), \(N\) is the mini-batch size, \(G\) is the number of groups chosen for the regularizer, and  \(\lambda_i\) are tunable coefficient that set the strength of sparsification.
\begin{equation}
\small
\label{eq:optimization}
\min_{\bm{W}} \left(\frac{1}{N} \sum_{i=1}^{N}  l(y_i, f(x_i, \bm{W})) + \sum_{g=1}^{G} \lambda_g\cdot||\bm{W}_{g}||_2\right)
\end{equation}
This lasso regularization sparsifies groups of weights by forcing the weights in each group to very small values, when possible without incurring high error. After sparsification, we use a small threshold of \(10^{-4}\) to zero out these weights.

\medskip
\noindent\textbf{Proposed Group Lasso Design.}
We design a specific group lasso regularizer that groups the weights of each channel (input or output) of each layer. We also choose a single global regularization strength parameter \(\lambda\) rather than adjust the penalty per group. The resulting regularizer term is shown in \eq{eq:group_lasso}, where
\(L\) is the number of layers in the CNN and \(C_l\) and \(K_l\) are the number of input and output channels in a layer, respectively.
\begin{equation}
\small
\label{eq:group_lasso}
\lambda \cdot \sum_{l=1}^{L}  \bigg( \sum_{c_l=1}^{C_l} ||\bm{W}_{c_l,:,:,:}||_2  +  \sum_{k_l=1}^{K_l} ||\bm{W}_{:,k_l,:,:}||_2 \bigg)
\end{equation}
Prior work proposes to penalize each channel proportionally to its number of weights in order to maintain similar regularization strength across all channels~\cite{simon2012standardization,alvarez2016learning}. Instead, we choose to use a single global regularization penalty coefficient because this emphasizes reducing computation over reducing model size. All convolution layers of a CNN have similar computation cost. Because early layers have fewer channels and each channel has larger features, each channel of their layers involves more computation. Therefore, applying a single global penalty coefficient effectively prioritizes sparsifying large features, which leads to greater computation cost reduction. We do not apply group lasso to the input channels of the first convolution layer and the output neurons of the last fully-connected layer, because the input data and output predictions of a CNN have logical significance and should always be dense.

\medskip
\noindent\textbf{Regularization Penalty Coefficient Setup.}
\label{subsec:reg_coeff}
To use lasso regularization from the beginning of training, the penalty coefficient \(\lambda\) should be carefully set to both maintain high prediction accuracy and to achieve a high pruning rate. We develop a new technique to set this strength coefficient without requiring resource-intensive hyper-parameter tuning. 
To do so, we choose $\lambda$ using the ratio of group lasso regularization loss out of the total loss (the sum of the group lasso regularization loss and the classification loss). This \textit{group lasso penalty ratio} is shown in \mbox{\eq{eq:lasso_ratio}}.
Based on our observations of several CNN models (ResNet32/50 and VGG11/13) and training data (CIFAR10, CIFAR100, and ImageNet), we find that using a group lasso penalty ratio of 20-25\% robustly achieves high structural model pruning (> 50\%) with small accuracy impact (< 2\%).

\begin{equation}
\small
\label{eq:lasso_ratio}
 Lasso\;penalty\;ratio = \frac{\lambda\sum_{g}^{G}||\bm{W}_{g,:}||}{l(y_i, f(x_i, \bm{W})) + \lambda\sum_{g}^{G}||\bm{W}_{g,:}||}
\end{equation}
We compute this using the random values to which weights are initialized at the beginning of training and the cross-entropy loss calculated after the very first network forward propagation. This penalty coefficient is set once at the first training iteration and maintained through training. Without our approach, prior work searches for a desired lasso regularization penalty coefficient, e.g., by trying random coefficient values until one that has a small impact on accuracy is found~\cite{alvarez2017compression,NIPS2016_6504}. This can potentially require many training runs for each CNN being trained and increase total training time.

\begin{figure}[h]
    \centering
    \includegraphics[width=0.48\textwidth]{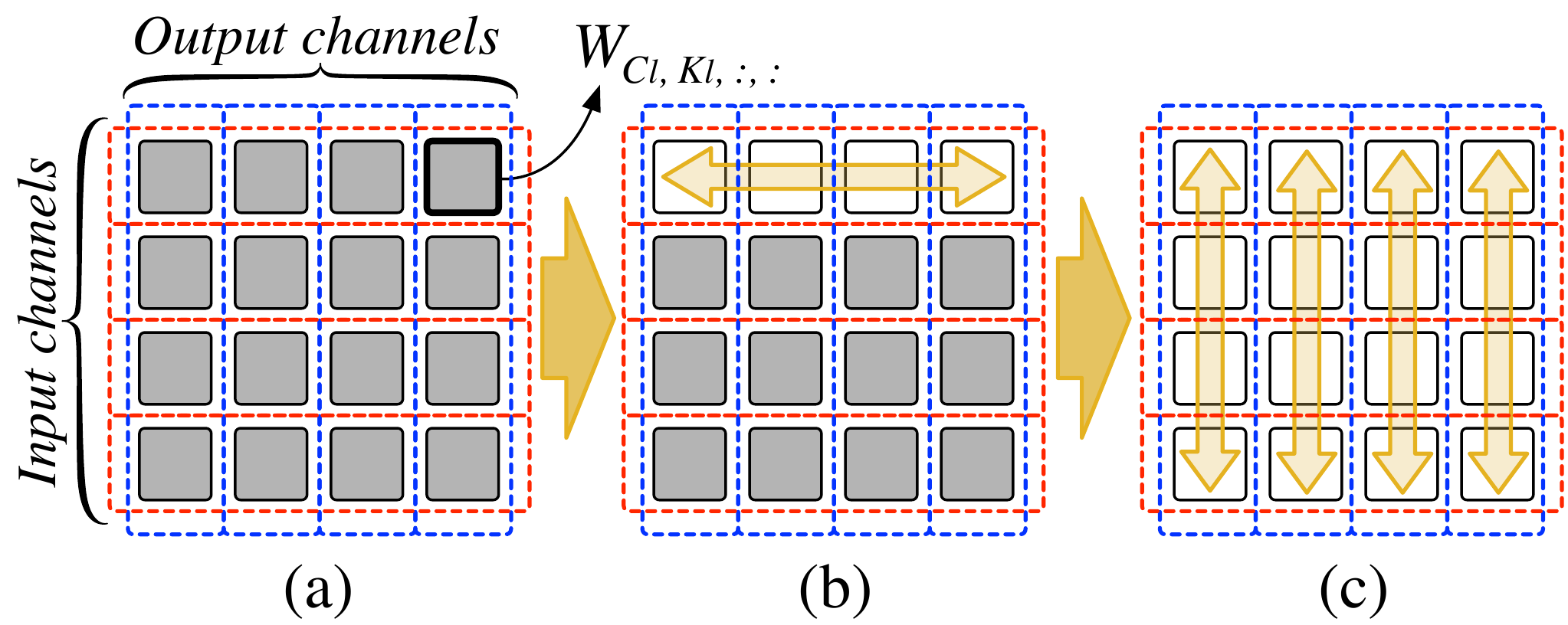}
    \vspace*{-6mm}
    \caption{Group lasso regularization structure of a convolution layer: Weights of a filter (each square box) affect the sparsification of weights in both input and output channels (red and blue dotted boxes). The white filters are zeroed-out after sparsification.}
    \label{fig:lasso_overlap}
\end{figure}

\medskip
\noindent\textbf{Layer Removal by Overlapping Regularization Groups.}
Wen et al.~\cite{NIPS2016_6504} propose to use layer-wise lasso groups for regularization in order to remove layers of a CNN with short-cut connections. However, we do not include such grouping in our regularizer. We find that because there is an overlap in the weights between input and output channel lasso groups (\fig{fig:lasso_overlap}{a}), unimportant layers are eventually removed even without additional layer-wise weight regularization. As an example, when an input channel becomes sparse (\fig{fig:lasso_overlap}{b}) by lasso regularization, it gradually sparsifies all the intersecting output channels (c), eventually leading to the entire layer to become zero.

\subsection{Dynamic Network Reconfiguration}
\label{subsec:net_reconf}

The main goal of \squeezetrain is reducing the training cost and time by continuously pruning the spasified channels or layers and reconfiguring the network architecture into a more cost-efficient form during training. There are two main concerns with doing so. The first is that pruning while training might prematurely remove weights that are unimportant early in training but become important as training proceeds. The second, is that the overhead of processing a pruned network exceeds any benefits realized by training a smaller model.

\medskip
\noindent\textbf{Early Weight Pruning.}
A prior pruning mechanism for CNNs that uses group lasso regularization, \emph{SSL}~\cite{NIPS2016_6504}, maintains the sparsified channels until the end of training instead of removing them from the model. This is because pruning while training prohibits weights from ``reviving'' and becoming non-zero as training proceeds. This can happen as gradients flow back from the last FC layer and potentially increase the value of previously-zeroed weights. However, we observe that already-zeroed input and output channels of convolution layers are likely to suppress such revived weights from ever becoming large. This can be inferred from the equation of the local weight gradients for a layer \(l\):
\begin{equation}
\small
\label{eq:weight_grad}
\frac{\partial L}{\partial \bm{W}_{l}} = \bm{z}_{l-1} \circledast \frac{\partial L}{\partial \bm{x}_l}^T
\end{equation}
Here, \(\circledast\) is convolution operator, and \(z_{l-1}\) and \(\frac{\partial L}{\partial \bm{x}_l}\) are the input activations (or input features) and the upstream gradients from the subsequent normalization layer. If a channel is sparsified and zeroed-out, its convolution outputs \(x_{l-1}\) are zeroed and they remain zero after normalization and activation layers, meaning that \(z_{l-1}\) is zero. Also, if an input channel of the subsequent convolution layer (\(l\)) is zeroed, the upstream gradients of this input channel are forced to be small. Thus, the gradients after passing the normalization layer \(\frac{\partial L}{\partial x_l}\) are also kept small by the gradient equation from~\cite{ioffe2015batch}. Therefore, using \eq{eq:weight_grad}, the gradients of zeroed weights are forced to remain very small and often zero, effectively restricting the previously zeroed weights from reviving.

\begin{figure}[t!]
    \centering
    \includegraphics[width=0.48\textwidth]{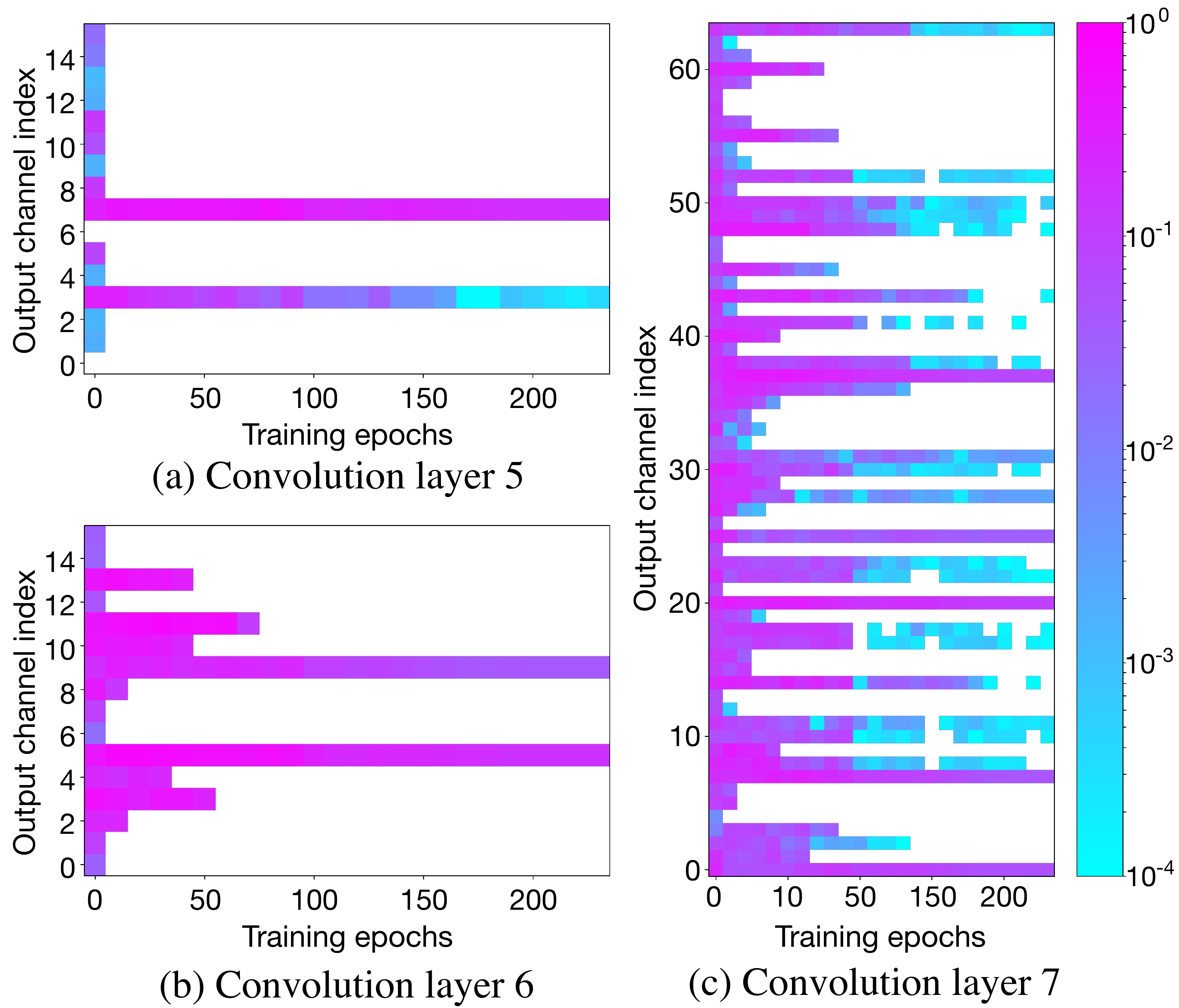}
    \vspace*{-5mm}
    \caption{The maximum absolute weight value of each output channel over training epochs. Three convolution layers belong to one residual path of ResNet50 trained on CIFAR10.}
    \label{fig:sparsity_pattern}
\end{figure}

This behavior is apparent in \fig{fig:sparsity_pattern} that shows the output channel sparsity of three layers of ResNet50~\cite{he2016deep} trained on CIFAR10 dataset across training epochs. Each point in the graph is the absolute maximum value among the parameters of each output channel. If the absolute maximum value of a channel becomes smaller than the threshold (\(10^{-4}\)), the parameters of the channel are zeroed out (white). Convolution layers 5 and 6 are typical and none of the weights from the zeroed output channels revive. Although some parameters in output channels of convolution layer 7 revive, their weight values are still very small and near the threshold, indicating a very small contribution to the prediction accuracy of the final learned model. Similar patterns are observed in all convolution layers of different ResNet and VGG models on CIFAR10/100, with all layers exhibiting no significant revived parameters.


\medskip
\noindent\textbf{Robustness to Reconfiguration Interval.}
\label{subsec:channel_union}
We now discuss the practical mechanisms for performing dynamic reconfiguration. We define a reconfiguration interval, such that after every such interval the zeroed input and output channels are pruned out. Note that if all the sparsified input and output channels are pruned, there is a possibility of a mismatch between the dimensions of the output channels of one layer and to the input channels of the next. To maintain dimension consistency, we only prune the intersection of the sparsified channels of any two adjacent layers. At any reconfiguration, all training variables of the remaining channels (e.g., parameter momentums) are kept as is.

The reconfiguration interval is the only additional hyperparameter added by \squeezetrain. Intuitively, a very short reconfiguration interval may degrade learning quality while a long interval offers less speedup opportunity. We extensively evaluate the impact of the reconfiguration interval in \sect{subsec:eval_main} and show that training is robust within a wide range of reconfiguration intervals.

\begin{figure}[t!]
    \centering                                                                            
    \subfloat[Residual modules: the channel dimensionality of the convolution layers sharing the same node (\ding{182}, \ding{183}, \ding{184}, and \ding{185}) should match.]{
        \includegraphics[width=0.46\textwidth]{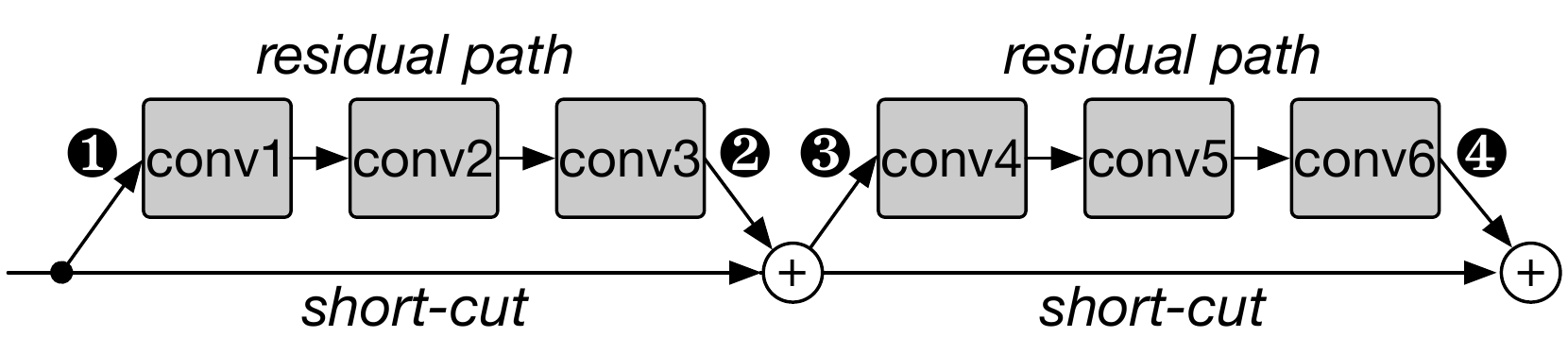}
        \label{fig:channel_base}
    }\\
    \vspace*{-1mm}
    \subfloat[Channel gating: \textit{channel select} and and \textit{channel scatter} layers match the channels indexes.]{
        \includegraphics[width=0.46\textwidth]{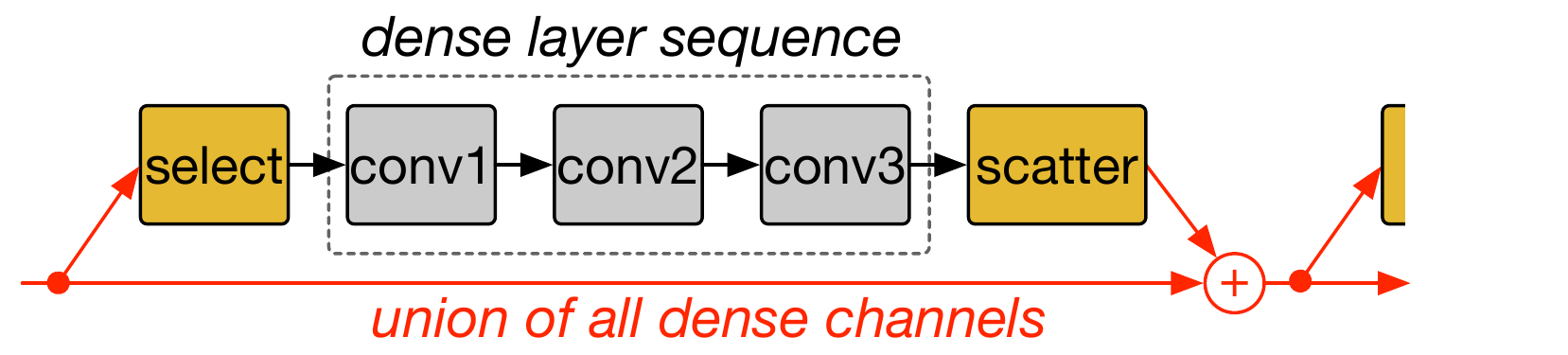}
        \label{fig:channel_gating}
    }\\
    \vspace*{-1mm}
    \subfloat[Channel union: the first and the last convolution layers of each residual path contain sparse channels.]{
        \includegraphics[width=0.46\textwidth]{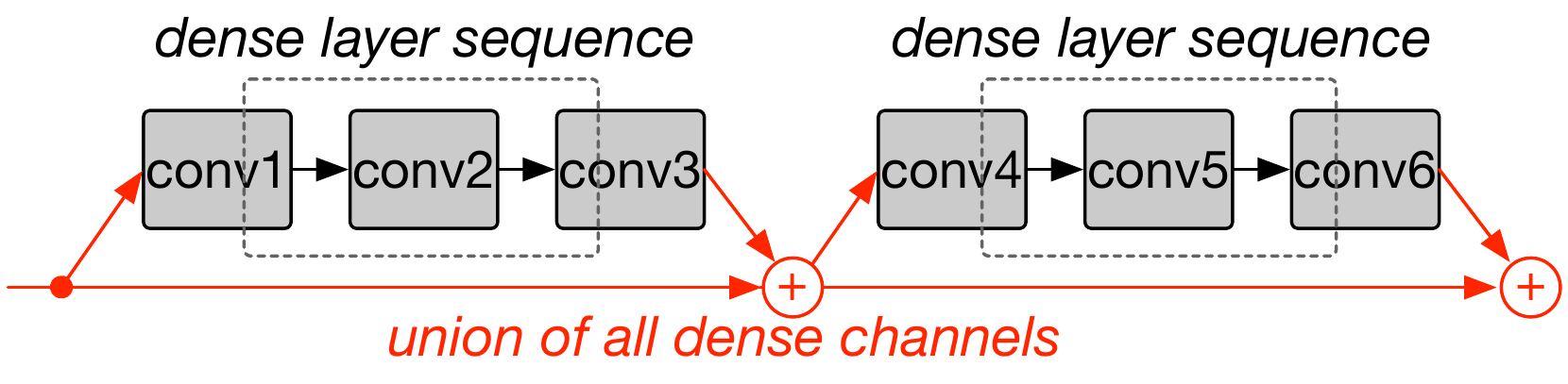}
        \label{fig:channel_union}
    }
    \caption{Channel indexing for CNNs with short-cut paths.}
    \label{fig:channel_indexing}
\end{figure}

\medskip
\noindent\textbf{Channel Union for CNNs with Short-cut Connections.}
Short-cut connections are widely adopted in modern CNNs, including ResNet and its many variations~\cite{he2016identity,xie2017aggregated,hu2017squeeze,huang2017densely}. They enable deep networks by mitigating the vanishing-gradients problem and achieve high accuracy~\cite{he2016deep}. For such CNNs, the channels of the convolution layers at a merge-point should match in dimensionality after each reconfiguration for proper feature propagation (\fig{fig:channel_base}). We propose two mechanisms to ensure this occurs. The first is {\bf channel gating} layers that add gating to each residual branch to match dimensions, as shown in \fig{fig:channel_gating}. This ensures that all convolution layers in a residual block operate only on dense channels by gathering and scattering the dense channel indices. This improves on the channel sub-sampling approach proposed by~\cite{he2017channel}, with channel sub-sampling only avoiding redundant computation of the very first convolution layer of each residual block.

We evaluate channel gating on an NVIDIA V100 GPU and find that channel gating involves significant memory accesses for tensor reshaping needed for channel indexing that often slows down training. Therefore, as an alternative, we propose {\bf channel union} that does not need any tensor reshaping and data indexing. Channel union prunes only the intersection of sparsified channels of all neighboring convolution layers within a residual stage (residual blocks sharing the same node). For instance, in \fig{fig:channel_union}, the union of the dense input channels of convolution layer 1 and 4 and the dense output channels of convolution layer 3 and 6 are maintained. As each following residual path adds new information to the shared node, the early convolution layers in the stage (convolution layer 1) have to process operations from the sparse channels, thereby performing redundant operations.

\begin{figure}[t!]
    \centering
    \includegraphics[width=0.48\textwidth]{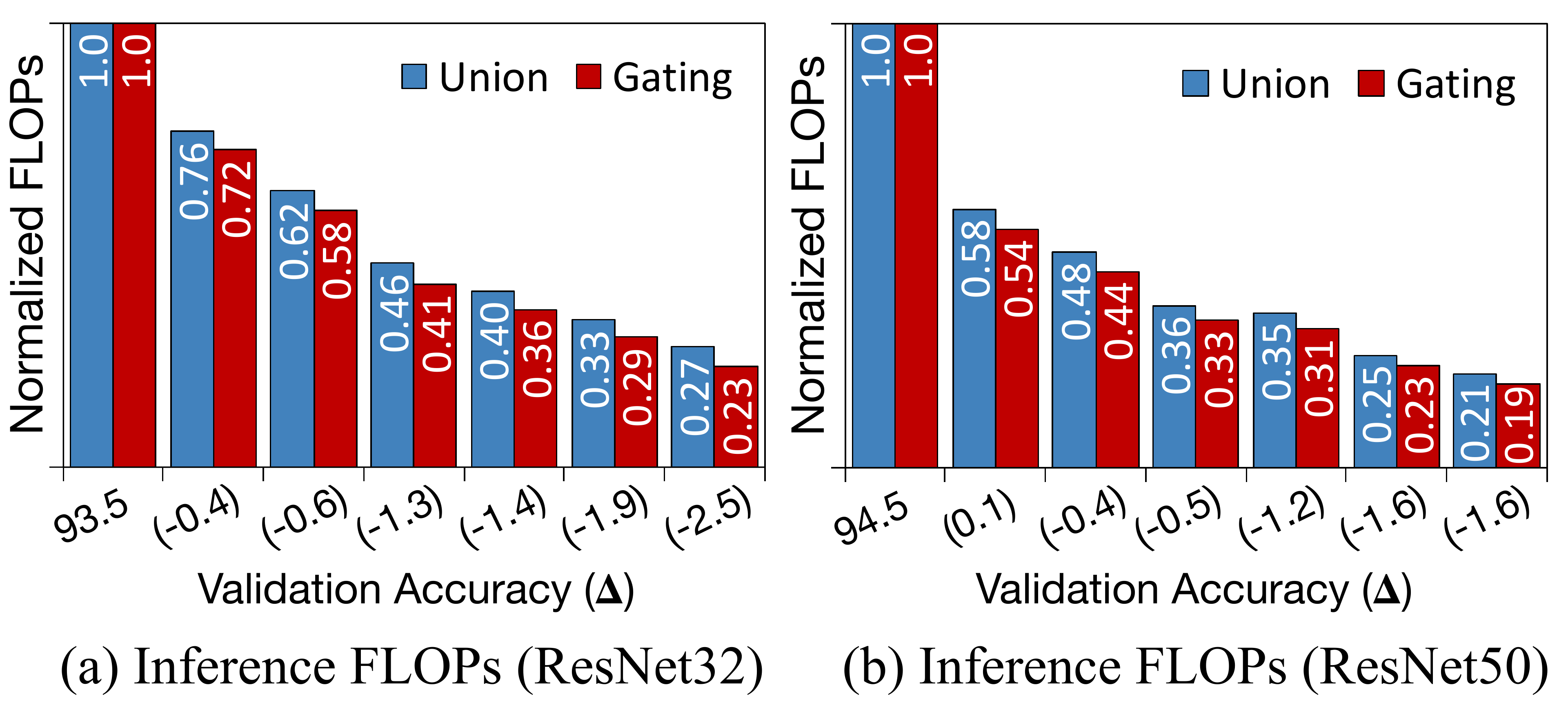}
    \vspace*{-5mm}
    \caption{Normalized training and inference FLOPs of ResNet32 and ResNet50 on CIFAR10 with different pruning intensity.}
    \label{fig:gating_vs_union}
  \vspace*{-2mm}
\end{figure}

\begin{figure}[t!]
    \centering
    \includegraphics[width=0.47\textwidth]{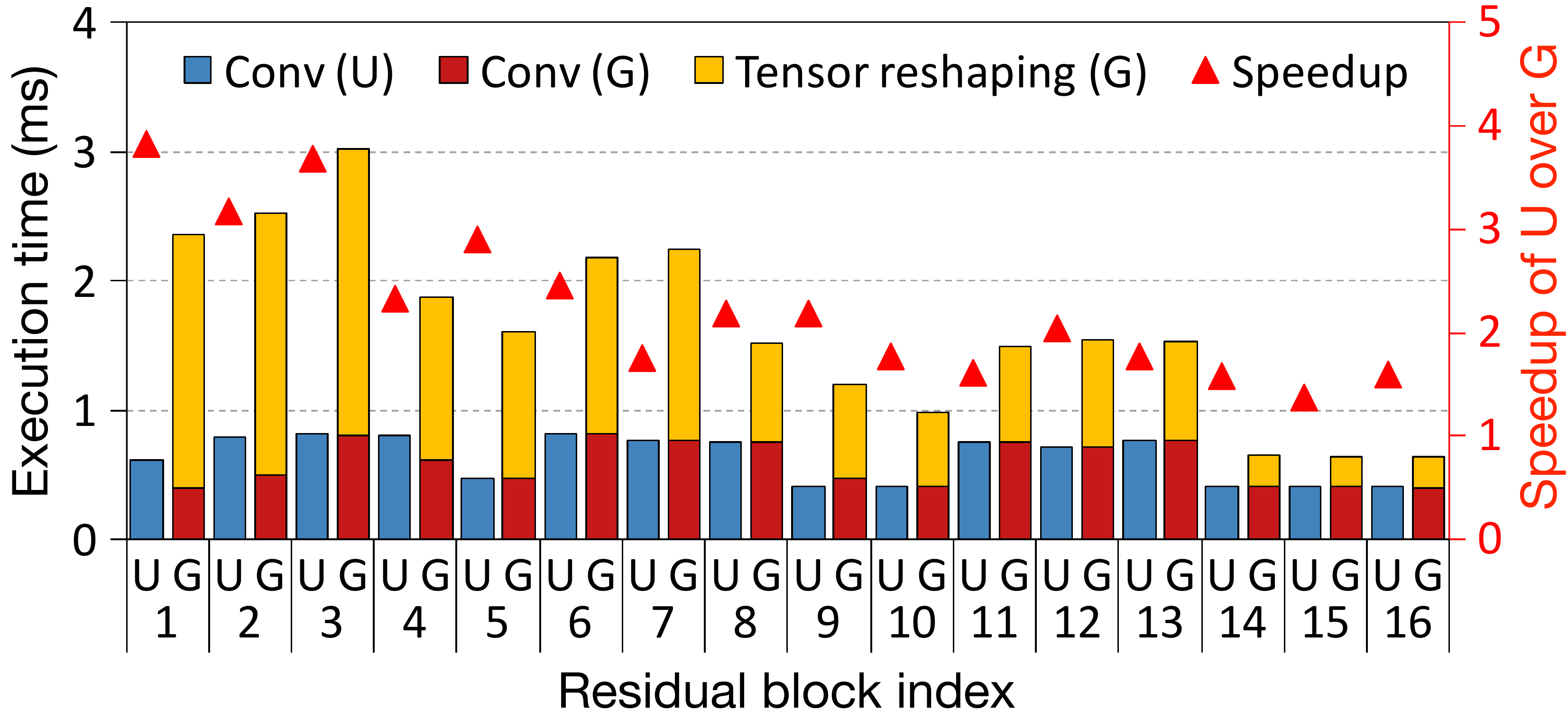}
    \vspace*{-2mm}
    \caption{Per-layer execution time of channel gating and channel union for ResNet50 for ImageNet. G and U indicate channel gating and channel union respectively.}
    \label{fig:gating_time}
  \vspace*{-2mm}
\end{figure}

However, our experiments show that the additional FLOPs from channel union, as compared to channel gating are very small. \fig{fig:gating_vs_union} compares the normalized inference FLOPs of channel gating and channel union for ResNet32 and ResNet50 pruned with different intensities. Across different pruning rates, the FLOPs difference is only 1-6\%, but the overhead saved from indexing is substantial. Additionally, this FLOPs difference does not grow with increasing layer depth as shown in \fig{fig:gating_vs_union} comparing ResNet32 and ResNet50. \fig{fig:gating_time} shows the measured per-layer (the last layer of each residual block) execution time of ResNet50 for ImageNet. For all residual blocks, channel union shows far less execution time compared to channel gating. Especially, the tensor reshaping time of early layers has bigger overhead as their activation size is eight times bigger than the layers in the last residual block.


\subsection{Dynamic Mini-batch Adjustment}
\label{subsec:dma}
As discussed in \sect{sec:cnn_train}, training with a large mini-batch reduces the frequency of costly inter-GPU communication and off-chip memory accesses for model updates. In addition, a larger mini-batch increases the data parallelism available at each network layer improving HW utilization.
We find that our gradual channel and layer pruning simultaneously reduces available data parallelism and decreases the memory capacity requirement for training, thus allowing the use of larger mini-batches. The latter allows us to compensate for the former as follows.

We propose \emph{dynamic mini-batch adjustment} to increase the size of the mini-batch by monitoring the memory context volume of a training iteration, which is gradually decreased by \squeezetrain. When channels are pruned by \squeezetrain, the output features corresponding to these channels are also not generated, which reduces the off-chip memory space required for back-propagation. In particular, early layers of a CNN have larger features and removing the channels of these layers effectively reduces the training memory requirement. Using a global regularization penalty, \squeezetrain prunes the channels of early layers by a larger ratio than prior work and enables using a larger mini-batch over training epochs. At every network architecture reconfiguration, \squeezetrain monitors the off-chip memory capacity required for a training iteration and increases the mini-batch size when possible.

However, dynamically increasing the size of the mini-batch alone does not guarantee high prediction accuracy as it is the hyperparameter closely coupled with the learning rate. To maintain the algorithmic functionality, we increase the learning rate by the same ratio of a mini-batch size increase to maintain the same learning quality. This mechanism is similar to adjusting the mini-batch size instead of decaying the learning rate, as proposed by Smith et al.~\cite{smith2017don}. However, our proposed mechanism is different in that we change the mini-batch size and learning rate dynamically at any point during training, unlike this prior work that changes them at the original learning rate decay points. Note that dynamic mini-batch size adjustment relies on the linear relation between mini-batch size and learning rate. For other deep learning applications that have a different relation, an appropriate learning rate adjustment rule can be adopted instead (e.g., the square root scaling rule for language models~\mbox{\cite{puri2018large}}). We evaluate dynamic mini-batch adjustment by training ResNet50 on both CIFAR and ImageNet datasets and confirm that it maintains equally high accuracy compared to the baseline \squeezetrain.

\medskip\noindent 
The overall PruneTrain training flow is summarized in \algo{algo:squeezetrain}.
\vspace*{-3mm}
\begin{algorithm}[h]
\caption{\squeezetrain neural network model training flow}
\label{algo:squeezetrain}
\begin{algorithmic}[1]

\LineComment{B: training dataset, M: mini-batch}
\LineComment{\(\bm{W_i}\): weights of a network model at {\em{i}} th iteration}
\LineComment{Net: network architecture}
\LineComment{LR: learning rate}
\For {\(e \gets 0, training\_epochs\)}
    \LineComment{Mini-batch iterations over the training dataset}
    \For {\(n \gets 0, \ceil{ \frac{B_{size}}{M_{size}} } \)}
        \State \(i = n + \big( e \times \ceil{ \frac{B_{size}}{M_{size}}} \big) \)
        \State \(loss_{1}, features = {\bf{ForwardProp}}(M_n, \bm{W_i})\)
        \State \(loss_{2} = {\bf{GroupLassoReg}}(\bm{W_i})\)

        \LineComment{Set the group lasso regularization penalty coefficient}
        \If {i = 0}
            \State \(\lambda\) = {\bf{SetCoeff}}\((loss_{1}, loss_{2})\)
        \EndIf
        \State \(loss\) = \(loss_{1}\) + \(\lambda \times loss_{2}\)

        \LineComment{Process network back-propagation and model updates}
        \State \(\bm{\Delta W_i}\) = {\bf{BackProp}}(\(loss\), \;\(features\), \;\(\bm{W_i}\))
        \State \(\bm{W_{i+1}}\) = {\bf{Optimizer}}(\(\bm{W_i}\), \;\(\bm{\Delta W_i}\), \;LR)
    \EndFor

    \If {{\bf{IsReconfigurationInterval}}(\(e\))}
        \LineComment{Prune and reconfigure the network architecture}
        \State Net = {\bf{PruneAndReconfigNetwork}}(\(\bm{W_{i+1}}\), Net)
        \LineComment{Update the mini-batch size and LR}
        \State \(M_{size}\), \(LR\) = {\bf{UpdateMiniBatch}}(system\_memory, Net)
    \EndIf
\EndFor
\end{algorithmic}
\end{algorithm}
\vspace*{-2mm}


\section{Evaluation}
\label{sec:evaluation}

\noindent\textbf{Evaluation Methodology.}
We evaluate \squeezetrain on both small (CIFAR10 and CIFAR100~\cite{krizhevsky2009learning}) and large datasets (ImageNet~\cite{imagenet_cvpr09}). We train four CNNs (ResNet32, ResNet50, VGG11, and VGG13) on CIFAR and ResNet50 on ImageNet, which is the most commonly used modern CNN for modern vision applications~\cite{lin2017focal,he2017mask,lin2017feature}. We use a mini-batch size of 128 and 256 (64 per GPU) for CIFAR and ImageNet training runs and a learning rate of 0.1 for both as the baseline hyperparameters~\cite{he2016deep}. We use four NVIDIA 1080 Ti and V100~\cite{volta2017whitepaper} GPUs for ImageNet training and a single TITAN Xp GPU~\cite{pascal2017whitepaper} for CIFAR training. We build \squeezetrain using PyTorch~\cite{paszke2017automatic}. Because of limited resources, we perform sensitivity evaluation primarily with CIFAR and evaluate functionality and final efficiency with ImageNet.

\subsection{Model Pruning and Training Acceleration}
\label{subsec:eval_main}

We first present our evaluation results on CIFAR and ImageNet in \tab{tab:compare_main}. We report 4 metrics: the training and inference FLOPs (FP operations), measured training time, and validation accuracy. 
Training time does not include network architecture reconfiguration time, which we do optimize and occurs only once in many epochs.
We compare the training results of ResNet and VGG using \squeezetrain with the dense baseline. We use the same number of training iterations for both the dense baseline and \squeezetrain to show the actual training time saved by \squeezetrain. 
We use 182 epochs~\cite{he2016deep} and 90 epochs to train CNNs on CIFAR and ImageNet, respectively.

\begin{table}[t!]
    \caption{Training FLOPs and time compared to the dense baseline: top1 validation accuracy of the dense baselines for CIFAR10: ResNet32 (93.6), ResNet50 (94.2), VGG11 (92.1), VGG13 (93.9), and for CIFAR100: ResNet32 (71.0), ResNet50 (73.1), VGG11 (70.6), VGG13 (74.1), and for ImageNet: ResNet50 (76.2)}
    \vspace*{-1.0mm}
    \centering
    \noindent\resizebox{1.0\linewidth}{!}{
        \tabulinesep=0.8mm
        \renewcommand{\arraystretch}{0.7}
        \begin{tabu}{ccccccc}
            \midhline
                       Dataset & Model & \makecell{Val. Accuracy \(\Delta\) \\ (fine-tunning)}  & \makecell{Train. FLOPs \\ (time)} & \makecell{\small{Inf.}\\FLOPs} \tabularnewline
            \midhline
            \multirow{4}{*}{CIFAR10}  & ResNet32 & -1.8\% & 47\% (81\%) & 34\% \tabularnewline
                                      & ResNet50 & -1.1\% & 50\% (81\%) & 30\% \tabularnewline
                                      & VGG11    & -0.7\% & 43\% (57\%) & 35\% \tabularnewline
                                      & VGG13    & -0.6\% & 44\% (57\%) & 37\% \tabularnewline
            \hline
            \multirow{4}{*}{CIFAR100} & ResNet32 & -1.4\% & 68\% (88\%) & 54\% \tabularnewline
                                      & ResNet50 & -0.7\% & 47\% (66\%) & 31\% \tabularnewline
                                      & VGG11    & -1.3\% & 53\% (74\%) & 43\% \tabularnewline
                                      & VGG13    & -1.1\% & 58\% (67\%) & 48\% \tabularnewline
            \hline
            \multirow{3}{*}{ImageNet} & \multirow{3}{*}{ResNet50} & -1.87\% (-1.58\%) & 60\% (71\%, *66\%) & 47\% \tabularnewline
                                      &                           & -1.47\% (-1.16\%) & 70\% (76\%, *72\%) & 56\% \tabularnewline
                                      &                           & -0.24\% (+0.20\%) & 97\% (98\%, *98\%) & 88\% \tabularnewline
            \midhline

        \end{tabu}
    }
    \begin{flushleft}
    \small{* Measured using V100 GPUs}
    \end{flushleft}
    \vspace*{-2mm}
    \label{tab:compare_main}
\end{table}

For ResNet32 and ResNet50 on CIFAR10, \squeezetrain reduces the training FLOPs by $\sim$50\% with a minor accuracy drop compared to the dense baseline. The compressed models after training show only 34\% and 30\% of the dense baseline inference cost for ResNet32 and ResNet50, respectively. The results of ResNet32/50 on CIFAR100 show similar patterns, which exhibits the robustness of \squeezetrain, given that CIFAR100 is a more difficult classification problem. For CIFAR100, \squeezetrain reduces the training and inference FLOPs by 32\% and 46\% for ResNet32, and 53\% and 69\% for ResNet50, while losing only 1.4\% and 0.7\% of validation accuracy, respectively compared to the dense baseline. These results show that \squeezetrain reduces more training FLOPs from a deeper CNN model, since more unimportant channels and layers are sparsified and removed early in the training. \squeezetrain also achieves high model compression with similar validation accuracy loss for both VGG models on CIFAR.

PruneTrain also shows high training cost savings for ResNet50 trained on ImageNet: 40\%, 30\%, and 3\% for three different pruning strengths (0.25, 0.2, and 0.1). Thus, we conclude that PruneTrain is robust to changes in CNN model and dataset complexity. The trained ResNet50 shows 53\%, 44\%, and 12\% reduced inference FLOPs with 1.87\%, 1.47\%, and 0.24\% accuracy loss, respectively. In addition, with extra training epochs for fine-tuning without group lasso regularization, we could recover 0.3\% additional accuracy for the regularization strengths of 0.25 and 0.2, and achieve even better accuracy than the baseline by 0.2\% for the regularization strength of 0.1. Although not shown in the table, PruneTrain also saves 37\%, 33\%, and 5\% of off-chip memory accesses of BN (batch normalization) layers for ResNet50 with the three different regularization strengths. Since the performance of BN layers is bounded by memory access bandwidth, reducing their memory traffic has a significant impact on the overall CNN model training time.

\begin{table}[t!]
    \caption{Inference performance comparison (number of images per second and relative speedup by PruneTrain). The three ResNet50 results on ImageNet use different regularization strengths of 0.25, 0.2, and 0.1.}
    \vspace*{-1.0mm}
    \centering
    \noindent\resizebox{1.01\linewidth}{!}{
        \tabulinesep=0.8mm
        \renewcommand{\arraystretch}{0.7}
        \begin{tabu}{cccccc}  
            \midhline
            \multirow{2}{*}{Dataset} & \multirow{2}{*}{Model} & \multicolumn{2}{c}{Batch size=10}  & \multicolumn{2}{c}{Batch size=100}  \tabularnewline
                                     &                        & Base & PruneTrain                        & Base & PruneTrain \tabularnewline
            \midhline
            \multirow{4}{*}{CIFAR100} & ResNet32               & 3038 & 4081 (1.34$\times$)                   & 18587 & 24759 (1.33$\times$) \tabularnewline
                                      & ResNet50               & 1442 & 1442 (1.18$\times$)                   & 7847  & 11865 (1.51$\times$) \tabularnewline
                                      & VGG11                  & 5534 & 5534 (1.44$\times$)                   & 15489 & 23878 (1.54$\times$) \tabularnewline
                                      & VGG13                  & 5197 & 5197 (1.38$\times$)                   & 12845 & 21075 (1.64$\times$) \tabularnewline
            \midhline
            \multirow{3}{*}{ImageNet} & \multirow{3}{*}{ResNet50} & \multirow{3}{*}{610} & 937 (1.53$\times$)  & \multirow{3}{*}{772}  & 1194 (1.55$\times$) \tabularnewline
                                      &                           &                      & 833 (1.36$\times$)  &                       & 1047 (1.36$\times$) \tabularnewline
                                      &                           &                      & 661 (1.08$\times$)  &                       & 813 (1.05$\times$) \tabularnewline
            \midhline
        \end{tabu}
    }
    \vspace*{-2mm}
    \label{tab:inference_comp}
\end{table}

\begin{figure*}
    \centering
    \subfloat{
        \includegraphics[width=0.44\textwidth]{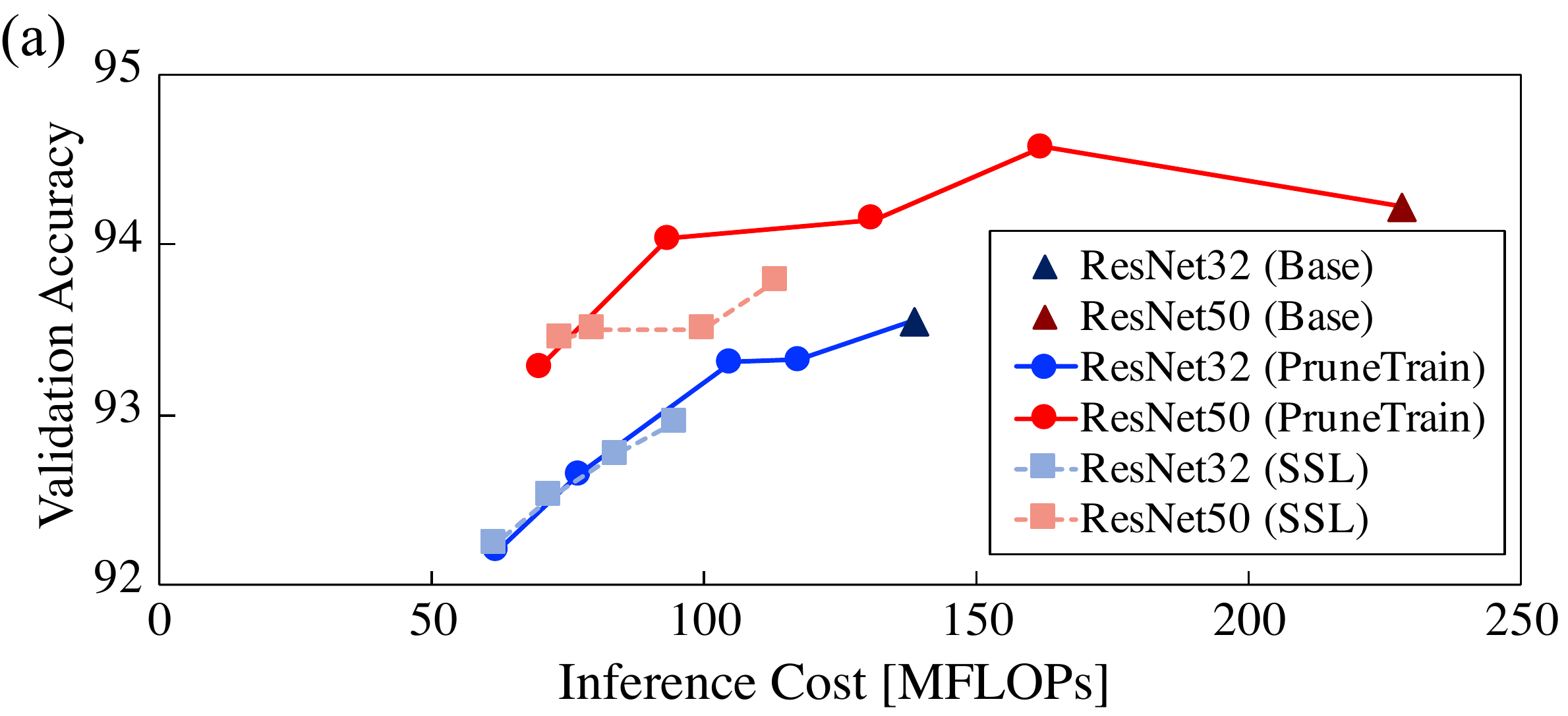}
        \label{fig:comp_to_priorwork_cifar10}
    }\hspace*{2mm}
    \subfloat{
        \includegraphics[width=0.535\textwidth]{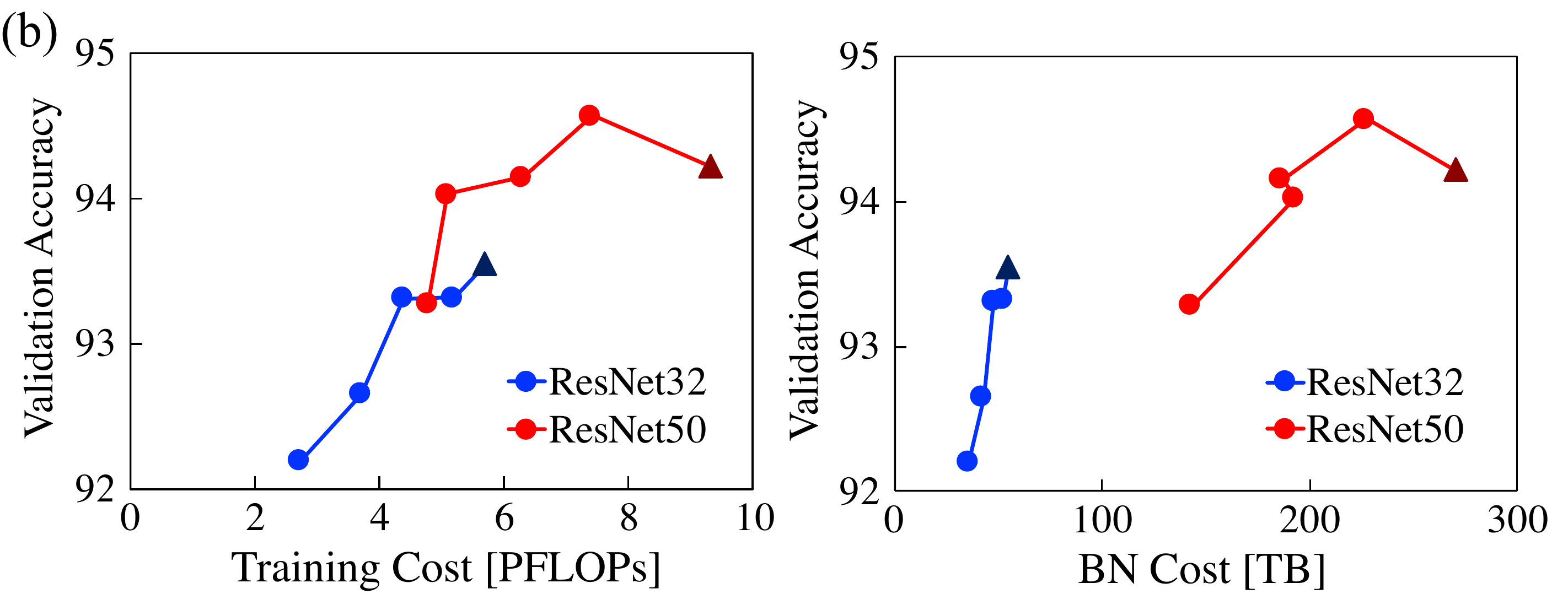}
        \label{fig:comp_to_priorwork_cifar10_train}
    }\\\vspace*{-3mm}
    \subfloat{
        \includegraphics[width=0.44\textwidth]{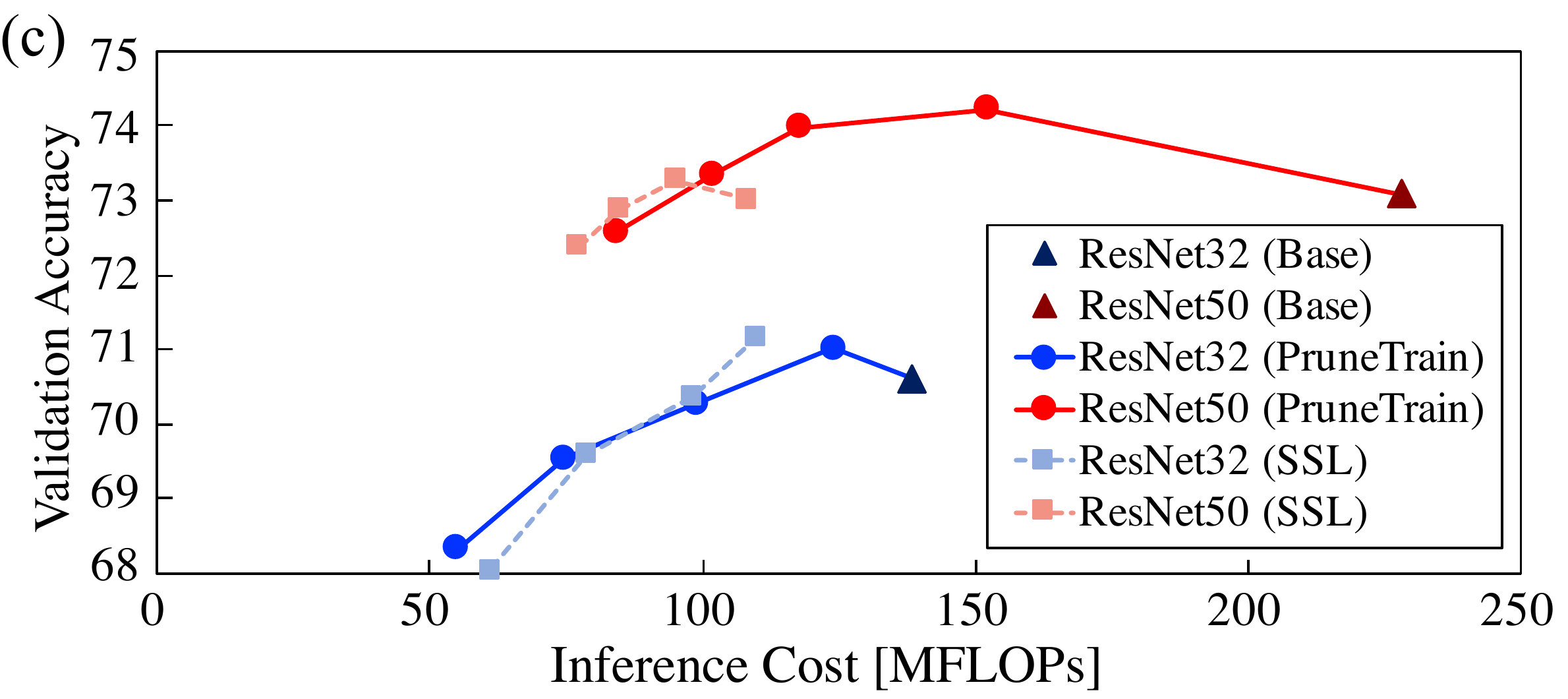}
        \label{fig:comp_to_priorwork_cifar100}
    }\hspace*{2mm}
    \subfloat{
        \includegraphics[width=0.535\textwidth]{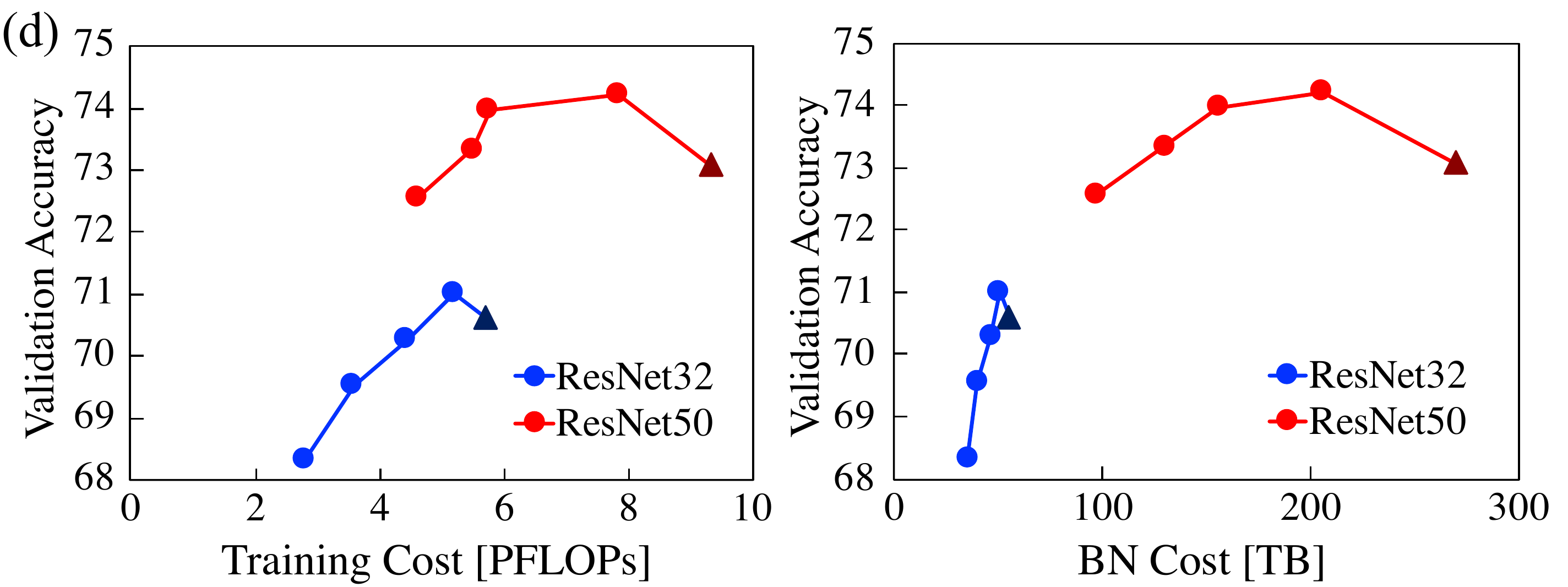}
        \label{fig:comp_to_priorwork_cifar100_train}
    }
    \vspace*{-2mm}
    \caption{(a) Inference FLOPs and the validation accuracy by different regularization ratios of \squeezetrain and SSL for ResNet32/50 on CIFAR10 (c) and on CIFAR100, (b) Training FLOPs and BN cost by accuracy of \squeezetrain for ResNet32/50 on CIFAR10 and on (d) CIFAR100. (The triangles in all figures represent the dense baseline)}
    \label{fig:comp_to_priorwork}
\end{figure*}

The measured training time reduction is smaller compared to the saved training FLOPs across datasets and CNN models. This is mainly caused by the reduced data parallelism at each layer after pruning, which decreases GPU execution resource utilization. Also, SIMD utilization within the GPU cores decreases for some layers due to the irregular channel dimensions after pruning and reconfiguration. In particular, for CIFAR10 and and CIFAR100, ResNets shows lower training time saving compared to VGGs, because it has many layers with reduced parallelism. In comparison, VGG has fewer layers with wider data parallelism and utilization is impacted less by pruning. For ImageNet, the training time saving of ResNet50 is bigger when V100 GPUs are used. This is because high off-chip memory bandwidth of V100~\mbox{\cite{jedec2016hbm2}} makes the execution time portion of memory bandwidth-bound layers smaller, which eventually makes the training time saving by the pruned computations more visible in the overall training time.

We also compare the performance of the trained models in terms of inference images per second (\mbox{\tab{tab:inference_comp}}). We evaluate using two different batch sizes of 10 and 100 using mixed precision~\mbox{\cite{micikevicius2017mixed}}, which we execute on one TITAN Xp GPU. Overall speedup of PruneTrain is lower than the saved inference FLOPs in \mbox{\tab{tab:compare_main}} because of resource underutilization. Therefore, processing 100 samples shows performance that is equal to or slightly better than the batch size of 10. Also, since ResNet50 for ImageNet has more channels, its PruneTrain inference performance is better than the CNN models for CIFAR100 given the ratio of their pruned FLOPs.  

\subsection{Comparison to Prior Work}
\label{subsec:eval_main}

\noindent\textbf{Comparison to Pruning From a Pre-trained Model (SSL).}
We verify that pruning while training from scratch shows comparable compression quality and accuracy as following the current best practice of training from a pre-trained model as done by SSL~\cite{NIPS2016_6504}. The comparison results are summarized in \fig{fig:comp_to_priorwork}, which plots the tradeoffs between both inference and training cost and validation accuracy for ResNet32/50 on CIFAR10/100. We sweep the group lasso penalty ratio from 0.05 to 0.2 with an interval of 0.05. Since Wen et al.~\cite{NIPS2016_6504} do not discuss how to set the group lasso penalty coefficient, we apply our proposed mechanism to SSL as well.

Results for inference (\fig{fig:comp_to_priorwork_cifar10} and \fig{fig:comp_to_priorwork_cifar100}) demonstrate that \squeezetrain is, in fact, superior to pruning from a pre-trained model. We make three important observations. First, for ResNet50, \squeezetrain attains higher accuracy than the baseline dense model while still reducing cost. Accuracy is highest at around 150 MFLOPs/inference compared to the dense 230 MFLOPs/inference. We attribute this to the regularizer we use for pruning also leading to better generalization~\cite{yuan2006model}. Second, \squeezetrain and SSL achieve comparable accuracy-cost tradeoffs, yet \squeezetrain offers a wider tradeoff range. Third, pruning is a very effective way to learn a good CNN model---starting from the complex ResNet50, \squeezetrain is able to learn a network model that is simultaneously more accurate and lower-cost to use.

\fig{fig:comp_to_priorwork_cifar10_train} and \fig{fig:comp_to_priorwork_cifar100_train} show the training-cost tradeoff curve. We do not show the training cost of SSL, because its training protocol first trains the dense network and then prunes, resulting in a cost that's almost 3 times higher than baseline. We show two aspects of training cost: the computation required for training and the memory traffic needed for the bandwidth-limited batch-normalization layers (bandwidth has lower impact on other layers). \squeezetrain reduces both the computation and memory traffic with a minor accuracy loss compared to the dense baseline (triangles in the graph). The shape of computation tradeoff curve is similar to that of inference. Because \squeezetrain gradually and continuously prunes the network to reduce its training cost over time, it can start from the complex ResNet50 and learn a better model in less training time compared to conventional dense ResNet32 training. Interestingly, unlike FLOPs, the memory traffic does not scale as well with regularization strength. This is because the regularization learns a different number of channels for different layers and the per-channel computation and memory cost are not always correlated; e.g., removing a channel of a 1x1 convolution layer decreases computations less than that of a 3x3 convolution layer, but their memory cost reduction for batch normalization is the same.

\begin{table}[!b]
    \caption{Comparison to AMC (Auto ML for Model Compression): compression results of ResNet56 on CIFAR10. The results of AMC are taken directly from~\cite{he2018amc}.}
    \vspace*{-1.0mm}
    \centering
    \noindent\resizebox{0.96\linewidth}{!}{
        \tabulinesep=0.8mm
        \renewcommand{\arraystretch}{0.7}
        \begin{tabu}{ccccc}
            \midhline
            Method & \makecell{Base Val.\\accuracy} & \makecell{Validation \\accuracy \(\Delta\)} & \makecell{\small{Inference}\\FLOPs} & \makecell{Removed\\layers}  \tabularnewline
            \midhline
            \squeezetrain  & 94.5\%    & -0.5\%    & 34\%   & 18 (21\%)\tabularnewline
            AMC            & 92.8\%    & -0.9\%    & 50\%   & Not known \tabularnewline
            \midhline
        \end{tabu}
    }
    \label{tab:compare_amc}
\end{table}

\medskip
\noindent\textbf{Comparison to Trial-and-Error Based Model Pruning.}
We compare the training results of \squeezetrain to AMC (Auto ML for model compression)~\cite{he2018amc} to show that learning the architecture by regularization during training leads to a better compression and accuracy tradeoff than trial-and-error based pruning from a pre-trained model (\tab{tab:compare_amc}). We use ResNet56 on CIFAR10 for comparison, which is the experimental setting used in AMC. While AMC reduces the inference FLOPs to 50\% with 0.9\% accuracy drop (after fine-tuning), \squeezetrain reduces an additional 16\% FLOPs while achieving higher accuracy by 0.4\%. While the capability of learning network depth was not discussed in AMC, \squeezetrain also learns depth and removes 21\% of the convolution layers of ResNet56. This layer removal is effective in reducing the actual inference latency because pruning layers does not decrease data parallelism and does not affect compute-resource utilization.

\subsection{Optimization and Sensitivity Evaluation}
\label{subsec:eval_main}

\begin{figure}[t!]
    \centering                                                                            
    \subfloat[ResNet50 on ImageNet: Memory requirement at every 5 epochs. The baseline mini-batch size per GPU of 64 is increased to 96 and 128 at the {\nth{10}} and {\nth{30}} epochs respectively, which fits the device memory capacity.]{
        \includegraphics[width=0.48\textwidth]{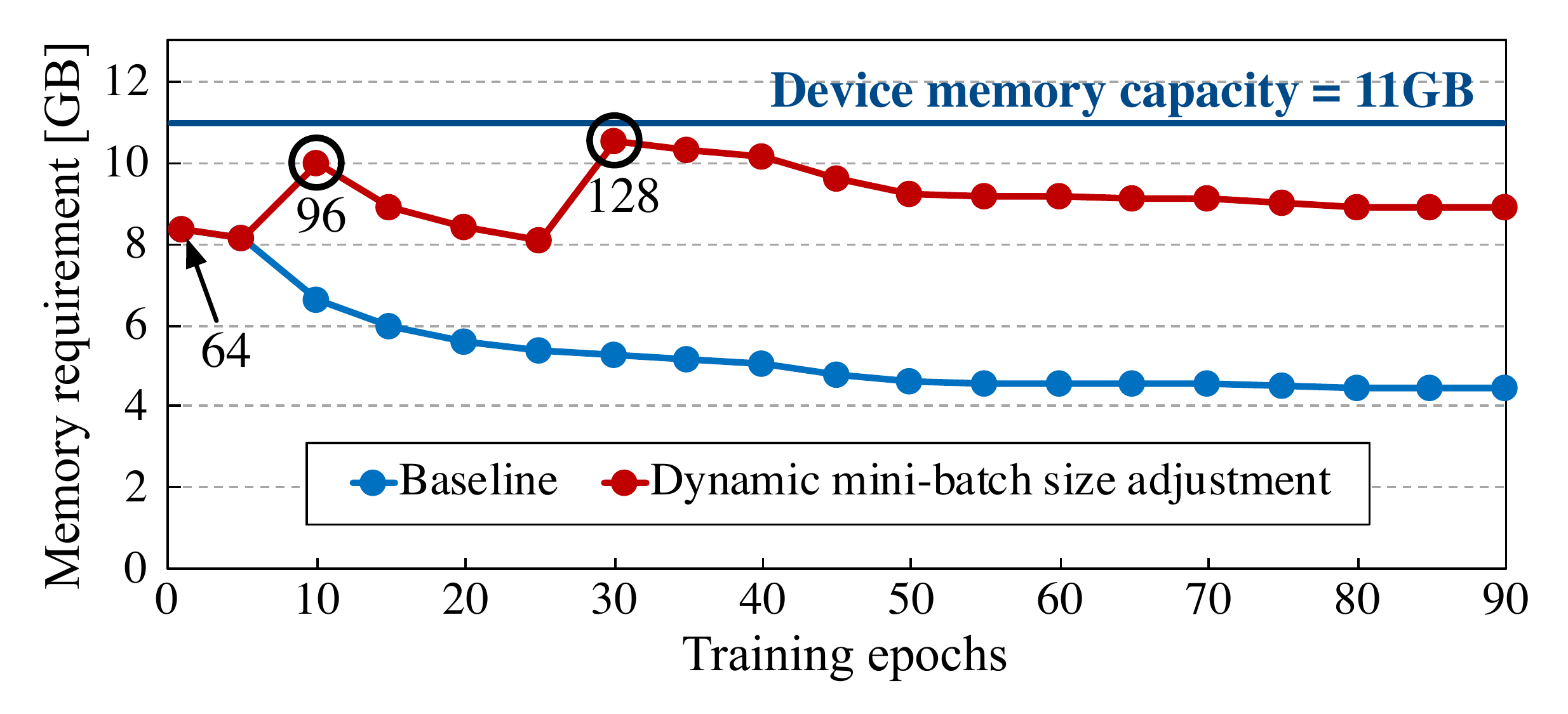}
        \label{fig:mem_capa_imagenet}
    }\\
    \subfloat[ResNet50 on CIFAR100: Normalized memory requirement every 10 epochs. The baseline mini-batch size of 128 is increased to 192, 224, 256, 288, and 320 at the {\nth{20}}, {\nth{30}}, {\nth{50}}, {\nth{70}}, and {\nth{120}} epochs respectively.]{
        \includegraphics[width=0.48\textwidth]{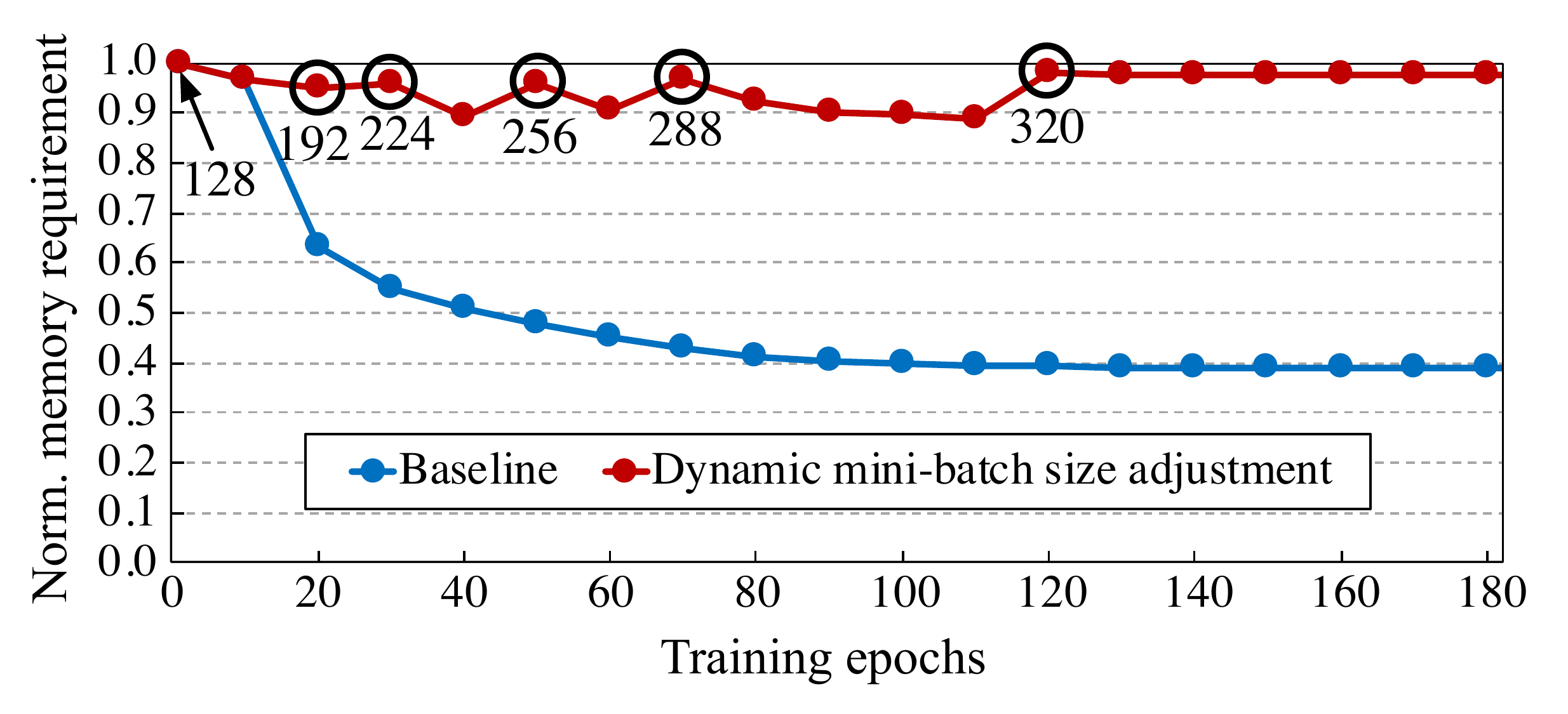}
        \label{fig:mem_capa_cifar100}
    }
    \vspace*{-1mm}
    \caption{Memory requirement of one training iteration per accelerator during training epochs.} \label{fig:mem_capa}
\end{figure}

\noindent\textbf{Dynamic Mini-Batch Size Adjustment.}
\fig{fig:mem_capa} shows the off-chip memory requirement per GPU for a single training iteration using \squeezetrain. We train ResNet50 for CIFAR100 and ImageNet datasets on a GPU with an 11 GB memory capacity (NVIDIA 1080 Ti). As training proceeds, the memory requirement gradually decreases due to pruning.

Once enough space is freed up, our proposed \textit{dynamic mini-batch size adjustment} mechanism increases the mini-batch size to fully utilize the off-chip memory capacity. As shown in \fig{fig:mem_capa_imagenet}, for ImageNet, we start with a per-GPU mini-batch of 64 (total of 256 across 4 GPUs), which is the largest mini-batch that can fit in the off-chip device memory. As the  memory requirement gradually decreases by pruning, we increase the per-GPU mini-batch from 64 to 96 and later to 128 at {\nth{10}} and {\nth{30}} epoch, respectively. The training context still fits in the GPU memory at each epoch. In this example, we use a mini-batch size adjustment granularity of 32 samples per GPU, but a smaller granularity can also be used.

The memory required by ResNet50 for CIFAR100 is already small. Hence, in order to demonstrate the effect of dynamic mini-batch size adjustment in this case, instead of trying to fit the largest mini-batch size possible in the GPU memory, we start with the standard mini-batch size of 128 (\fig{fig:mem_capa_cifar100}).

Then, as PruneTrain gradually reduces the memory requirement, we gradually increase the mini-batch size such that we maintain similar device memory capacity utilization. This is shown in \fig{fig:mem_capa_cifar100}, where we increase the mini-batch size by multiples of 32 up to, eventually, a mini-batch of 320, which is 2.5X larger than the initial mini-batch size. Note that increasing the mini-batch size not only increases the computational parallelism, it also linearly decreases the model update frequency. Reducing model update frequency can significantly accelerate distributed training by lowering inter-device communication and off-chip memory accesses.

\tab{tab:cifar100_table} compares the training time reduction with and without dynamic mini-batch size adjustment.
The table also compares the validation accuracy and final inference computation complexity in the two scenarios.
While dynamic mini-batch size adjustment barely affects the quality of learning and pruning, it has a high impact on training time. Dynamic mini-batch size adjustment improves accuracy by 0.3\% and raises the inference FLOPs by 3\% for CIFAR100 and reduces accuracy by 0.04\% and decreases the inference FLOPs by 1\% for ImageNet. It reduces the training time by 57\% and 34\% (39\% on a V100 GPU) compared to the dense baseline for CIFAR100 and ImageNet, respectively. This is also an improvement of 26\% and 17\% (14\% on a V100 GPU) compared to the naive PruneTrain for CIFAR100 and ImageNet, respectively.
Although the training time is substantially reduced, the impact is less than expected, given that we are enabling 2$\times$ more computational parallelism with fewer model updates than the naive \squeezetrain.
We suspect that this is caused by a sub-optimal GPU convolution kernel choice that comes from the increased data parallelism only in mini-batch dimension.

\begin{table}[h]
    \caption{Training time, inference FLOPs, and validation accuracy with and without dynamic mini-batch size adjustment for ResNet50.
    Top-1 validation accuracy of the dense baselines: ResNet50 trained on CIFAR100 (73.1) and on ImageNet (76.2).}
    \vspace*{-1.0mm}
    \centering
    \noindent\resizebox{1.0\linewidth}{!}{
        \tabulinesep=0.8mm
        \renewcommand{\arraystretch}{0.7}
        \begin{tabu}{cccccc}
            \thickhline
            Dataset                     & Model           & \makecell{Method} & \makecell{Train time\\reduction} & \makecell{Inference\\FLOPs} & \makecell{Val.\\Acc. \(\Delta\)} \tabularnewline
            \midhline
            \multirow{2}{*}{CIFAR100}   & \multirow{2}{*}{ResNet50}    & Naive      & 34\%  & 31\% & -0.7\%  \\
                                                                     & & Adjusted   & 43\%  & 34\% & -0.4\%  \tabularnewline
            \hline
            \multirow{2}{*}{ImageNet}   & \multirow{2}{*}{ResNet50}    & Naive      & 29\% (*34\%)  & 47.4\% & -1.87\%  \\
                                                                     & & Adjusted   & 34\% (*39\%)  & 46.4\% & -1.91\%  \tabularnewline
            \thickhline
        \end{tabu}
    }
    \begin{flushleft}
    \small{* Measured using V100 GPUs}
    \end{flushleft}
   \label{tab:cifar100_table}
\end{table}

\begin{figure}[h]
    \centering
    \includegraphics[width=0.48\textwidth]{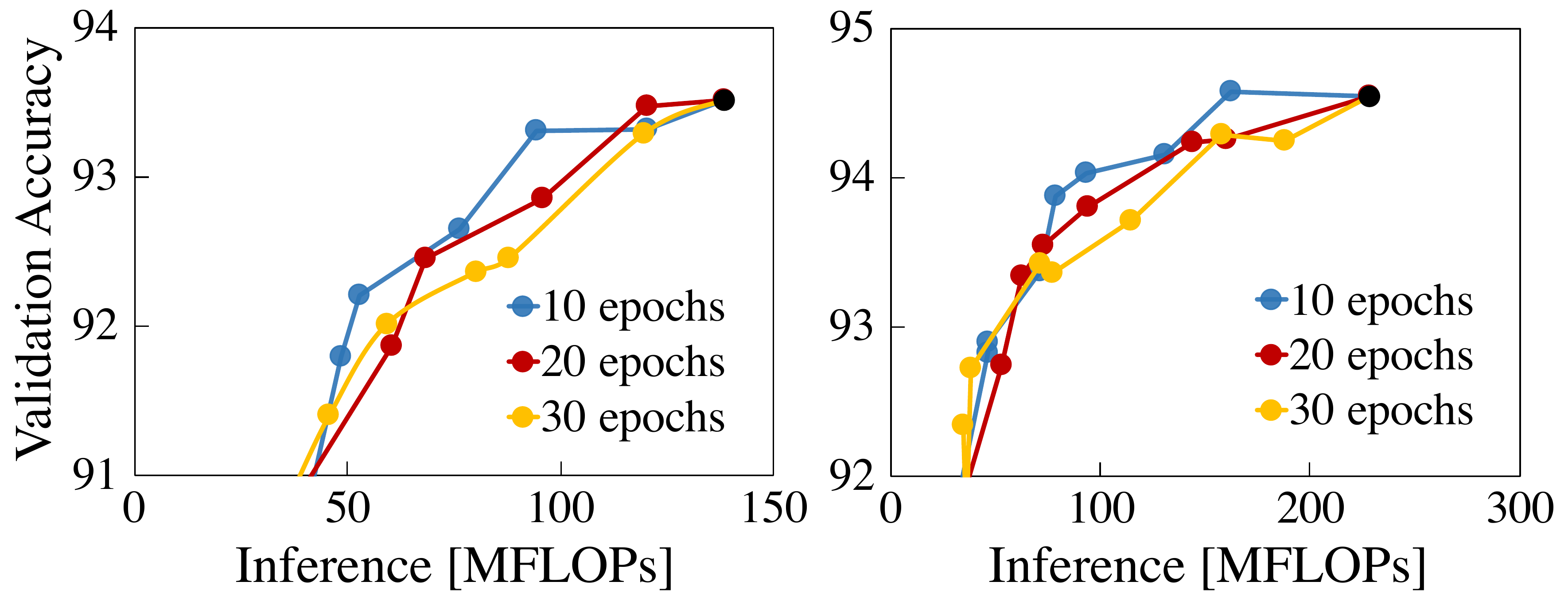}
    \vspace*{-5mm}
    \caption{Reduced inference FLOPs and validation accuracy by different network reconfiguration intervals. ResNet32 (Left) ResNet50 (Right) on CIFAR10.}
    \label{fig:interval_resnet}
\end{figure}

\medskip
\noindent\textbf{Network Reconfiguration Interval.}
\squeezetrain adds two hyper-parameters on top of dense training: sparsification strength, which we already discussed, and the reconfiguration interval. The reconfiguration interval affects training time by trading off the time overhead of manipulating the network model with greater savings of more-frequent pruning (actual removal of computation). The reconfiguration interval may also affect the compression and accuracy of the final learned model. Fortunately, the compression and accuracy achieved are insensitive to this hyper-parameter, as shown in \fig{fig:interval_resnet}, which shows the accuracy vs.~computation cost tradeoff curve for different intervals. Thus, the interval can be chosen to balance per-iteration performance gains with reconfiguration time overhead. The overhead depends on the specific framework used. We find that reconfiguring a network architecture every 10 epochs for CIFAR or 5 epochs for ImageNet has small overhead in our experiments.


%

\begin{figure}[!b]
    \centering
    \includegraphics[width=0.45\textwidth]{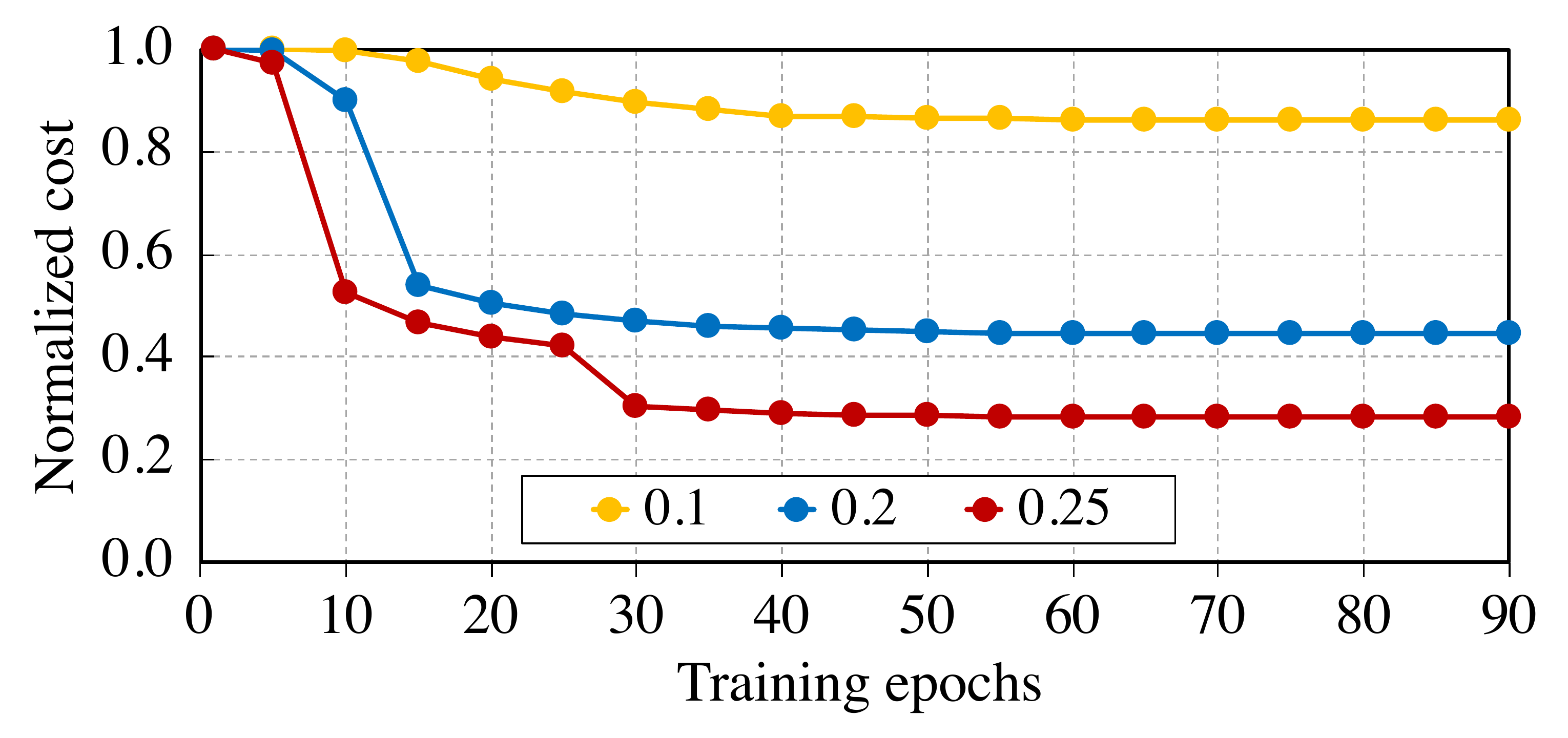}
    \vspace*{-3mm}
    \caption{Projected per-epoch communication cost of model updates based on hierarchical ring-allreduce. The communication cost is normalized to the cost of dense baseline ResNet50 training on ImageNet for different group lasso regularization penalty ratios.}
    \label{fig:imagenet_comm_cost}
\end{figure}

\noindent\textbf{Communication Cost Savings in Distributed Training.}
As training proceeds, the model size reduction by \squeezetrain leads to decreasing communication cost between GPUs. \fig{fig:imagenet_comm_cost} shows the projected decrease in communication cost during the training of ResNet50 for ImageNet. We model the communication cost using ring allreduce. The figure shows the communication cost per training epoch normalized to the dense baseline for different sparsification strengths (therefore, different pruning rates). Each time the network is reconfigured, the number of weights decreases, leading to a reduction in weight gradients communicated per training iteration. Furthermore, an aggressive sparsification strength (0.2 and 0.25) allows dynamic mini-batch adjustment to increase the mini-batch sizes (dotted lines), leading to further reduction in communication cost for later epochs. Overall, PruneTrian saves 55\% average communication cost regardless of the number of GPUs used for distributed training. This pruning-based communication reduction is orthogonal to other existing techniques for communication reduction in distributed training, e.g. weight gradient compression and efficient gradient reduction mechanisms, which can be used in conjunction with \squeezetrain for further communication improvements.

\begin{figure}[h]
    \centering
    \includegraphics[width=0.47\textwidth]{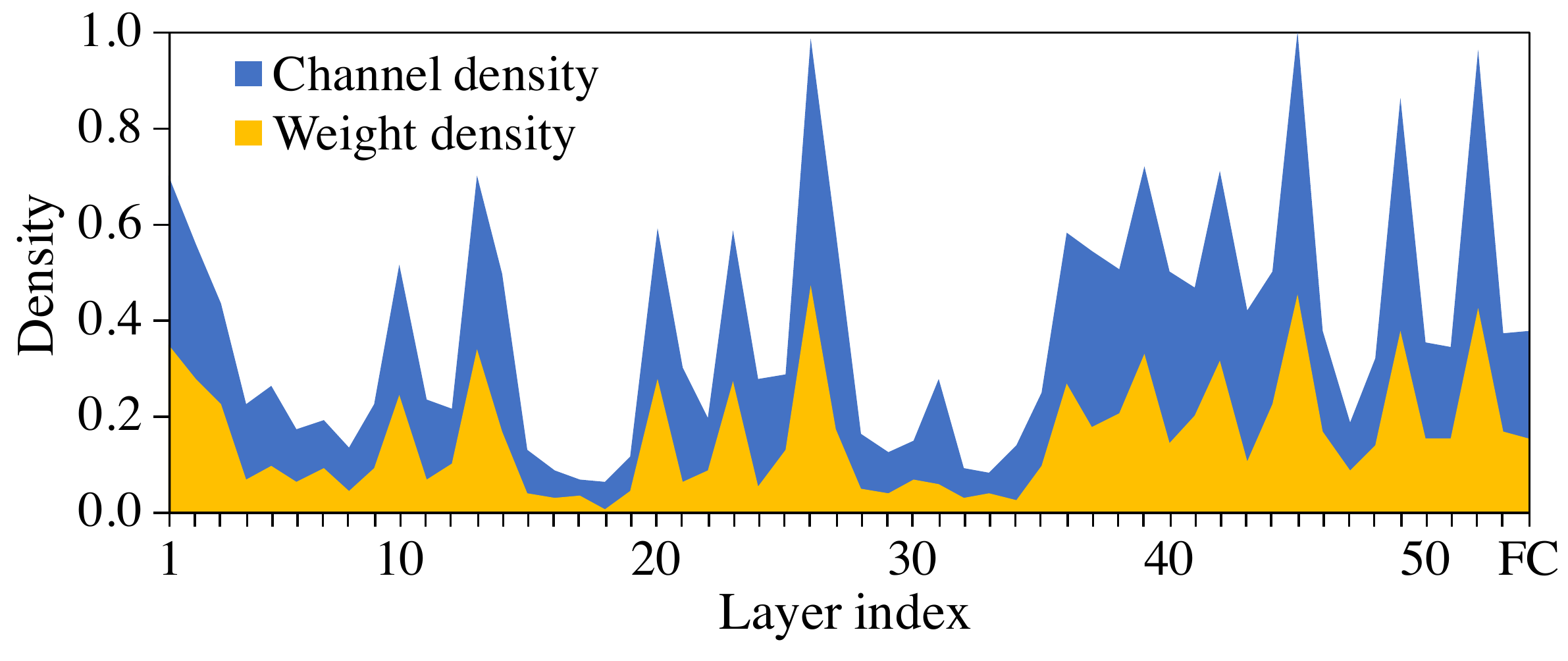}
    \vspace*{-2mm}
    \caption{Channel and weight density of each layer. (ResNet50 trained on ImageNet using \squeezetrain) The number in the x-axis indicates the convolution layer index.}
    \label{fig:imagenet_weight_density}
\end{figure}

\medskip
\noindent\textbf{Individual Weight Sparsity.}
\squeezetrain uses structured pruning of channels (and possibly layers) to learn a smaller, yet still dense model. This is important for high-performance execution on current hardware. However, the regularization leads to weight sparsity even within the remaining channels.  \fig{fig:imagenet_weight_density} shows the density of channels (input channel density $\times$ output channel density) and the density of weights for each layer in ResNet50 trained on ImageNet. Roughly half of all weights within the remaining channels (roughly half of all dense channels) are also near-zero and can be pruned. Such unstructured sparsity can be utilized to store the pruned model in a compressed form and to possibly further speed up execution if the inference hardware supports efficient sparse computations~\cite{han2016eie}.

\section{Conclusion}
\label{sec:conclusion}

In this paper, we propose \squeezetrain, a mechanism to accelerate the training from scratch of a network model, while pruning it for a faster inference.
\squeezetrain uses structural pruning, and continuously reconfigures the network architecture during training, so as to take advantage of the reduced model size not just during inference, but also during training.
This is based on our observation that while pruning with group lasso regularization, once a group of model parameters are forced to near-zero magnitude, they rarely revive during the rest of the training.
We propose three key optimizations for efficient implementation of \squeezetrain.
First, we update the group lasso regularization penalty coefficient such that we enable achieving high model pruning rate with minor accuracy loss during a single training run from scratch.
Second, we introduce channel union, a way to prune CNN models with short-cut connections to lower the overheads from naive channel indexing and tensor reshaping.
Lastly, we dynamically increase the mini-batch size while training with \squeezetrain, which increases the data parallelism and reduces communication frequency, leading to further training time saving.
Altogether, \squeezetrain cuts the computation cost of training modern CNNs (represented as ResNet50) at least by half, and up to 53\% and 40\% for small and large datasets, enabling 34\% and 39\% reduction in end-to-end training time respectively.


\section{acknowledgment}

The authors acknowledge Texas Advanced Computing Center (TACC) for providing GPU resources.

\balance
\bibliographystyle{ieeetr}
\bibliography{main.bbl}


\end{document}